\documentclass[lettersize,journal]{IEEEtran}

\usepackage{enumitem}
\usepackage{amsmath,amsfonts}
\usepackage{array}
\usepackage{color}
\usepackage{textcomp}
\usepackage{stfloats}
\usepackage{url}
\usepackage{verbatim}
\usepackage{graphicx}
\usepackage{cite}
\ifCLASSOPTIONcompsoc
  \usepackage[nocompress]{cite}
\else
  \usepackage{cite}
\fi

\ifCLASSINFOpdf
\else
\fi

\usepackage[numbers, sort&compress]{natbib}
\usepackage{amsmath, amssymb,upgreek}
\usepackage{algorithm}
\usepackage{algorithmic}

\usepackage{graphicx}
\usepackage{epstopdf}
\usepackage{makecell}
\usepackage{multirow, booktabs}
\usepackage{pifont}
\usepackage{ifthen}
\usepackage{hyperref}
\usepackage{cleveref}

\usepackage{soul}
\newtheorem{definition}{Definition}[section]
\newtheorem{theorem}{Theorem}[section]
\newtheorem{assumption}{Assumption}[section]

\newtheorem{lemma}[theorem]{Lemma}

\usepackage{array,graphicx,subfigure}

\usepackage{diagbox}
\begin{document}

\title{FairGFL: Privacy-Preserving Fairness-Aware Federated Learning with Overlapping Subgraphs}

\author{
Zihao Zhou, Shusen Yang, Fangyuan Zhao, Xuebin Ren 
\thanks{Received 15 May 2025; revised 5 November 2025; accepted 23 December 2025. This work was supported in part by Central Guidance for Local Science and Technology Development Fund Project under Grant 202507AC040003, in part by the National Key Research and Development Program of China under Grant 2022YFA1004100, and in part by the National Natural Science Foundation of China under Grants 62172329 and U21A6005. (Corresponding author: Shusen Yang.)}
\IEEEcompsocitemizethanks{\IEEEcompsocthanksitem Zihao Zhou is with the National Engineering Laboratory for Big Data Analytics (NEL-BDA) and School of Mathematics and Statistics, Xi'an Jiaotong University, Xi'an, Shaanxi 710049, China.\protect \
E-mails: wszzh15139520600@stu.xjtu.edu.cn.
\IEEEcompsocthanksitem Shusen Yang is with the National Engineering Laboratory for Big Data Analytics (NEL-BDA) and the Ministry of Education Key Lab for Intelligent Networks and Network Security, Xi'an Jiaotong University, Xi'an, Shaanxi 710049, China. \protect \
E-mail: shusenyang@mail.xjtu.edu.cn.
\IEEEcompsocthanksitem Fangyuan Zhao and Xuebin Ren are with the National Engineering Laboratory for Big Data Analytics (NEL-BDA) and School of Computer Science and Technology, Xi'an Jiaotong University, Xi'an, Shaanxi 710049, China.\protect \
E-mail: zfy1454236335@stu.xjtu.edu.cn, xuebinren@mail.xjtu.edu.cn. }  
\thanks{}}

\maketitle

\begin{abstract}
  Graph federated learning enables the collaborative extraction of high-order information from distributed subgraphs while preserving the privacy of raw data. However, graph data often exhibits overlap among different clients.
  Previous research has demonstrated certain benefits of overlapping data in mitigating data heterogeneity. However, the negative effects have not been explored, particularly in cases where the overlaps are imbalanced across clients. In this paper, we uncover the unfairness issue arising from imbalanced overlapping subgraphs through both empirical observations and theoretical reasoning.
  To address this issue, we propose FairGFL (FAIRness-aware subGraph Federated Learning), a novel algorithm that enhances cross-client fairness while maintaining model utility in a privacy-preserving manner.
  Specifically, FairGFL incorporates an interpretable weighted aggregation approach to enhance fairness across clients, leveraging privacy-preserving estimation of their overlapping ratios.  Furthermore, FairGFL improves the tradeoff between model utility and fairness by integrating a carefully crafted regularizer into the federated composite loss function.
  Through extensive experiments on four benchmark graph datasets, we demonstrate that FairGFL outperforms four representative baseline algorithms in terms of both model utility and fairness.
\end{abstract}


\section{Introduction}\label{sec:introduction}

\IEEEPARstart{G}{raph} data is widely used in various domains, such as neurosciences~\cite{vavsa2022null}, social networks~\cite{sharma2024survey} and financial transactions~\cite{zhang2023dynamic}, to model intricate relations among different entities.
For graph data mining, graph neural network (GNN)~\cite{kipfsemi, chen2021learning} is the state-of-the-art method, capable of extracting the structural features and higher-order interaction information among graph nodes.
Traditional GNN methods heavily rely on training on a large amount of graph data. However, due to privacy concerns and data isolation issues, it is challenging or even infeasible to gather sufficient graph samples, which are often distributed among different individuals or organizations.

Federated learning (FL)~\cite{mcmahan2017communication, huang2024federated, hanser2025data} is a promising paradigm for large-scale data analysis over distributed datasets. It enables the coordinated training of a global model without exposing the raw data of local clients. Due to the great potential, graph federated learning (GFL), which trains GNN models via FL, has also attracted significant interests from both academia~\cite{zhang2021subgraph, he2022spreadgnn, 10884817, 10858070} and industry~\cite{zhang2024federated, wu2022federated}. For example, Liu~\textit{et al.}~\cite{liu2022federated} propose a federated framework for social recommendation with GNN. Zhang~\textit{et al.}~\cite{zhang2023privacy} apply GFL for detecting financial frauds.

For GFL, the overlapping subgraphs among different FL clients are quite common in both academic research~\cite{zhang2021subgraph} and realistic scenarios, such as recommendation~\cite{wu2021fedgnn}, link prediction~\cite{baek2023personalized, zhang2021subgraph}.
For example, in the realm of recommendation, multiple e-commerce companies collaborate to train a recommendation model by bridging up their separate user-item subgraphs~\cite{wu2021fedgnn}. These subgraphs may share the same items, users (nodes), and their interactions (links). Another example pertains to the financial field. Multiple banks train a financial fraud detection model utilizing local bank transfer transaction graphs. Commonly, clients (nodes) register accounts across multiple banks and engage in transactions (links), hence resulting in overlaps within diverse graphs.

\textit{Some advantages of the overlapping data, such as the mitigation of data heterogeneity } \cite{zhao2018federated, han2021fedmes} have been explored in general FL on non-graph data (images, texts, etc.). For instance, Zhao~\textit{et al.}~\cite{zhao2018federated} share common data to reduce performance inconsistency across local models trained on heterogeneous data in federated image classification and keyword spotting tasks. Han~\textit{et al.}~\cite{han2021fedmes} leverage overlapping data to unify the training processes of image classification models built on non-IID data.

However, \emph{the adverse impact of overlapping data in FL, especially in GFL, remains unclear.} In this paper, we investigate this problem through extensive analytical experiments on benchmark datasets, including non-graph datasets (EMNIST ByClass, Sent140) and graph data (Cora, CiteSeer).
Our empirical observations reveal that imbalanced overlaps among clients may result in unfairness in the sense of significant model performance discrepancies, i.e., inconsistent model utility across distributed clients. Despite being not evident in FL for non-graph data, the unfairness is quite prevalent and severe in GFL. This distinction arises from disparity in node features and variations in neighboring information. In GFL, even completely overlapped subgraphs may have different links connected with other subgraphs, thus leading to significant heterogeneity and unfairness. Such unfairness would not only reduce the participation willingness of clients but also result in poor model generalization.

Recently, several cross-client fairness-aware methods~\cite{lifair, li2021ditto, cui2021addressing} have been proposed to improve the consistency of model performance across different clients in FL systems. Among those approaches, q-FedAvg~\cite{lifair} improves fairness by amplifying the loss value of disadvantaged clients, in which way their contributions to the model are enhanced and the performance disparity is mitigated.
Li~\textit{et al.}~\cite{li2021ditto} enhance fairness through a personalized algorithm that improves the utility of each local model while controlling its disparity with the global model.
Cui~\textit{et al.}~\cite{cui2021addressing} mitigate the performance inconsistency across local models by solving a min-max optimization problem that minimizes the maximum loss among them.
Without an in-depth understanding of the source, existing approaches address unfairness by enforcing the model performance consistency among clients in the model optimization process. Therefore, their effectiveness in addressing unfairness specifically sourced from imbalanced overlapping data is limited, as demonstrated in our experiments.

\emph{There exist two challenges} when mitigating the unfairness issue.
\emph{First}, it is difficult to design an interpretable algorithm to enhance fairness based on elusive imbalanced overlaps in GFL. One intuitive approach is to directly remove overlapping subgraphs, which, however, is impractical due to the inaccessible raw data in FL~\cite{10508131} and the diverse links connected with other subgraphs.
\emph{Second}, it is difficult to maintain the trade-off between fairness and model utility. To realize fairness, the model utility on some advantaged clients would decrease inevitably, which may lead to a significant decline in the overall model utility.

To address these challenges, we conduct the first systematic analysis of the impact of imbalanced overlapping subgraphs on fairness and propose FairGFL, a cross-client fairness-aware graph federated learning algorithm.
In short, our contributions can be summarized as follows.
\begin{itemize}[leftmargin = *]
  \item We provide comprehensive evidence to substantiate that imbalanced overlapping data across clients can result in severe unfairness, especially in GFL. Specifically, we first empirically observe the unfairness issue in extensive experiments on both benchmark non-graph datasets and graph data with imbalanced overlaps. Then, we theoretically reason about the unfairness issue by analyzing the difference in empirical loss functions between overlapping and non-overlapping settings.
  \item We propose FairGFL, a novel fairness-aware GFL algorithm to enhance fairness in a privacy-preserving manner and maintain model utility. In particular, we first introduce an evaluation metric of overlapping rate and apply a new LDP protocol to privately estimate the overlapping ratios of subgraphs among different clients. Based on the estimates, an interpretable weighted aggregation technique is then developed to enhance cross-client fairness, i.e. to achieve consistent performance across different clients. Finally, a well-crafted regularizer is incorporated into the federated composite losses to improve the tradeoff between the model utility and fairness.
  \item We evaluate FairGFL by conducting extensive experiments on four real-world graph datasets. The experimental results demonstrate that FairGFL outperforms four representative fair algorithms in terms of both fairness and model utility.
\end{itemize}

The rest of the paper is organized as follows. Section \ref{sec:relatedwork} reviews the related work. Section \ref{preliminary} presents the preliminary knowledge and some important definitions. Section \ref{sec:motivation} shows the empirical observations and the theoretical reasoning of unfairness issues caused by overlapping data. Section \ref{sec:method} describes the details of our solution. Corresponding theoretical analysis is provided in Section \ref{sec:analysis}. Section \ref{sec:experiments} shows the experimental results. Section \ref{sec:conclusion} concludes the paper.

\section{Related Work}\label{sec:relatedwork}

This section reviews the literature related to our work, with a specific focus on the studies on cross-client fairness, fair graph learning and graph federated learning.

\subsection{Cross-client Fairness}

A wide spectrum of fairness notions has been proposed in machine learning, including individual fairness, group fairness, and long-term fairness, each pursuing distinct objectives~\cite{zhang2023censored, hardt2016equality, du2021fairness}. For example, Group fairness ensures equitable outcomes for demographic groups~\cite{shen2022fair}, while long-term fairness promotes the sustained participation of underrepresented clients to mitigate global model bias~\cite{shi2023towards}.
In light of these considerations, we specifically focuses on cross-client fairness, which aims to ensure consistent model performance across all clients. Several methods have been proposed to address cross-client fairness in settings with heterogeneous data distributions. Li et al. introduced q-FedAvg~\cite{lifair}, which amplifies the influence of clients with higher losses; Li et al. proposed Ditto~\cite{li2021ditto}, balancing local performance with global utility; and Cui et al. developed FCFL~\cite{cui2021addressing}, formulating the problem as multi-objective optimization to mitigate performance inconsistency across clients.

However, none of these methods specifically target the unfairness issue arising from overlaps. Besides, they tend to enhance fairness at the expense of model utility and may fail in graph-specific tasks due to the neglect of graph structures. The complex interplay of diverse nodes and links inherently results in more pronounced and multifaceted data heterogeneity, thereby presenting unique challenges that conventional methods are ill-equipped to handle. In contrast, FairGFL provides the first in-depth understanding of the unfairness issue arising from imbalanced overlaps and utilizes novel interpretable techniques to mitigate the unfairness.

\subsection{Fair Graph Learning}
Most fair graph Learning (GL) studies focus on mitigating the performance inconsistency across models on different nodes (users) caused by some sensitive features, such as sex and race~\cite{ling2023learning, zhu2024one}. Nevertheless, disparities in other node features and variations in structural information can also lead to inconsistent performance~\cite{dong2021individual}. Dong et al.~\cite{dong2021individual} addressed this by enhancing individual fairness through minimizing cross-entropy loss between the ground truth and predicted similarity matrices of any two nodes' outputs. Furthermore, within GFL, a fairness notation involves higher contributions leading to increased rewards~\cite{pan2024towards}. Pan et al.~\cite{pan2024towards} achieved this by evaluating clients' contributions based on gradient similarity and data volume and then rewarding different clients through incentive mechanisms.

However, none of these studies investigate the model inconsistency issue in non-IID data settings in GFL. Additionally, existing fairness algorithms in GL can not be incorporated in GFL due to the nature of no access to raw data. In contrast, FairGFL first sheds light on the correlation between imbalanced overlapping subgraphs, data heterogeneity, and their detrimental effects on cross-client fairness.

\subsection{Graph Federated Learning}
Numerous studies~\cite{liu2024federated, fu2022federated} have been conducted on graph federated learning. We focus on three core challenges that are pertinent to this study, i.e. missing edges, non-IID data, and differential privacy.
First, in FL, some nodes on each client may have neighbors belonging to other clients~\cite{zhang2021subgraph, peng2022fedni}. In this scenario, any two clients may overlap in nodes but links~\cite{zhang2021subgraph}. To predict the missing links, Zhang \textit{et al.} optimized the loss function by jointly training a missing neighbor generation neural network.
Second, variations in graph structures across clients significantly impact the performance of GFL models. To address it, existing methods mainly focus on personalized learning~\cite{baek2023personalized, ye2023personalized} and knowledge sharing~\cite{tan2023federated}.
Third, privacy considerations extend to the safeguarding of both structural information and node features~\cite{lin2022towards, gauthier2023personalized}. For instance, Lin et al.~\cite{lin2022towards} applied differential privacy on nodes and neighboring matrix. To further improve the model utility, they make the matrix sparse based on $l_{1}$ regularization and the similarity of neighboring nodes.

However, existing research neglects the adverse impact of overlapping subgraphs on data heterogeneity and unfairness. Meanwhile, due to the privacy restriction of GFL, it is difficult to access the overlapping data across clients. This paper demonstrates the differentiated heterogeneous levels stemming from imbalanced overlapping subgraphs and proposes FairGFL to address it in a privacy-preserving manner.

\section{Preliminary}\label{preliminary}

\subsection{Fairness}

We consider the cross-client fairness which prioritizes the performance consistency across clients, similar to \cite{lifair, li2021ditto}. The definition is as follows.

\begin{definition}[Cross-client Fairness]
  Given a metric $\mathcal{F}$ that quantifies the performance consistency and a partial ordering relation $\preccurlyeq$ defined on the model parameters space. For the trained model $w$ and $\tilde{w}$, if the following condition holds
  \begin{equation}
    \mathcal{F}(w) \leq \mathcal{F}(\tilde{w}),
  \end{equation}
  we say $w \preccurlyeq \tilde{w}$, indicating that the model $w$ provides a fairer solution to the FL objective compared to the model $\tilde{w}$.

  If there exists a model $w$ such that, for any model $\tilde{w}$,
  \begin{IEEEeqnarray}{cl}
      w \preccurlyeq \tilde{w},
  \end{IEEEeqnarray}
  we say the model $w$ is a fair solution to the FL objective.
\end{definition}
There are several metrics available to measure performance consistency, such as variance, and the opposite number of entropy. Detailed formal expressions can be referred to \cite{lifair}. These commonly used metrics exhibit smooth characteristics.
Moreover, the partial ordering relation $\preccurlyeq$ possesses three desirable properties, i.e. reflexivity, antisymmetry, and transitivity.
Therefore, in the topologically bounded closed model parameters domain, a fair solution $w$ exists for the FL objective. This solution can achieve an optimal balance among different clients' performances.

\subsection{Local Differential Privacy Enhanced FL}
In FL, potential threats include security risks from attacks (e.g., Byzantine attacks or model inversion attack)~\cite{farhadkhani2022byzantine, 11196960} and privacy risks (inferring sensitive information) from malicious adversaries~\cite{zhu2019deep, shokri2017membership}. Robust aggregation and pruning methods are promising defenses against these attacks~\cite{farhadkhani2022byzantine, 11196960}. Besides, there exists other privacy-enhancing techniques to provide provable privacy guarantees, such as Local Differential Privacy (LDP). A randomized mechanism $\mathcal{M}$ satisfies $\epsilon$-LDP, w.r.t., for any input values $v, v_0 \in \mathcal{D}$ and any possible output $\mathcal{S} \subseteq \text{Range}(\mathcal{M})$, we have
\begin{IEEEeqnarray}{cl}
   \mathbb{P}[\mathcal{M}(v) \in \mathcal{S}] & \leq e^{\epsilon}  \mathbb{P}[\mathcal{M}(v_{0}) \in \mathcal{S}].
\end{IEEEeqnarray}
LDP involves a range of perturbation and sampling methods, such as random response, which are well-suited for various scenarios and types of private information.
For example, Wang~\textit{et al.}~\cite{wang2020federated} propose a novel Random Response with Priori (RRP) technique to protect local documents in a federated topic modeling task. 
Different from the existing gradient noising methods, we introduce a novel variant of the random response technique to sanitize the subgraphs for estimating the overlapping ratios to improve cross-client fairness\footnote{ The LDP protocol can be directly extended to protect the transmitted inner results, i.e., model updates~\cite{li2023multi}, for achieving more comprehensive protection. However, since privacy is not the main focus of this paper, FairGFL simply incorporates the LDP technique to protect the raw graph data from exposure.}.

\subsection{Graph Learning}

\begin{figure*}[!t]
  \centering
  \subfigure[Graph.]{
		\includegraphics[width=0.08\textwidth, trim={410 220 470 180}, clip]{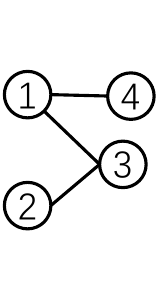}   
		\label{GFL:graph}
	}
  \hfil
  \subfigure[Graph learning.]{
		\includegraphics[width=0.4\textwidth, trim={192 155 261 170}, clip]{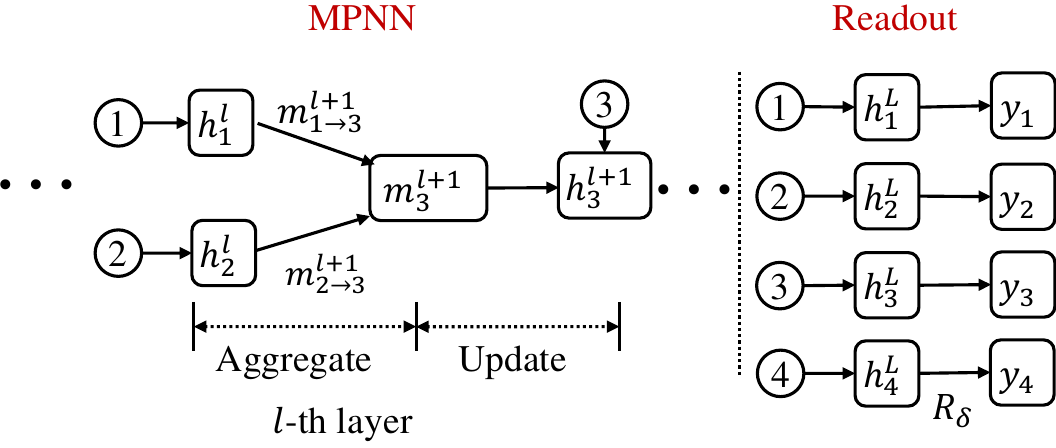}
		\label{GFL:GL}
	}
 \hfil
  \subfigure[Graph federated learning.]{
		\includegraphics[width=0.32\textwidth, trim={280 165 320 187}, clip]{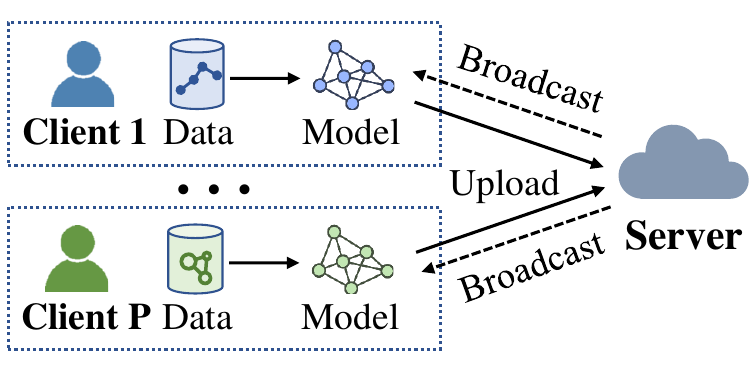}
		\label{GFL:GFL}
	}
  \caption{Workflow of node-classification graph learning and graph federated learning.}\label{GFL}
\end{figure*}

Graph learning encompasses three tasks: node classification task, graph classification task, and link prediction.
This paper specifically focuses on the node classification task.  Given a global graph denoted as $G = \{\mathcal{V}, E, \mathbf{X} \}$, where $\mathcal{V}$ represents the node set, $\mathbf{X}$ is the corresponding node feature set, and $E$ represents the link set, the node-classification graph learning network consists of two main phases, i.e., the message passing phase (implemented with a message-passing network, denoted MPNN) and the readout phase, as illustrated in \figurename~\ref{GFL:GL}.

The message passing phase contains two sub-procedures: (i) the model transforms and aggregates messages from neighboring nodes for each node, and (ii) the aggregated messages are further utilized to update the nodes' hidden states. Formally, the $l$-th layer of MPNN updates the hidden state of the $k$-th node with the following rule:
\begin{IEEEeqnarray}{lc}\label{graphlearningone}
  & m_{k}^{l+1} = \text{AGG}( \{  M_{\theta}^{l+1}(h_k^l, h_j^l, e_{k,j}), j \in \mathcal{N}_k  \} ),  \label{GL:eq1}  \\
  & h_{k}^{l+1} = U_{\phi}^{l+1} ( h_{k}^{l}, m_{k}^{l+1} ),  \label{GL:eq2}
\end{IEEEeqnarray}
where $h_{k}^{0} = x_{k}$ and $h_k^{l}$ represents the current hidden state, $e_{k,j}$ is the link features, $\mathcal{N}_{k}$ is the set of neighboring nodes, $\text{AGG}(\cdot)$ represents the aggregation function (e.g., SUM), and $M_{\theta}^{l+1} (\cdot)$ is the message generation function.
$U_{\phi}^{l+1} (\cdot)$ is the state update function that incorporates the aggregated feature $m_{k}^{l+1}$.

After the forward propagation of MPNN, the readout phase aggregates the outputs and infers predictions. Formally, it can be expressed as
\begin{IEEEeqnarray}{lc}\label{graphlearningtwo}
  & \hat{Y} = R_{\delta} ( \{ h_{k}^{L}, k \in [n] \} ),               \nonumber
\end{IEEEeqnarray}
where $n = |\mathcal{V}|$ represents the total number of graph nodes, $\hat{Y}=\{\hat{y}_{k}, k\in[n]\}$ represents the predictions, and $R_{\delta}$ represents the readout function. Based on predictions $\hat{Y}$, the model parameters $w$ are then updated by minimizing the following empirical risk function $F(w)$:
\begin{IEEEeqnarray}{lc}\label{GLRisk}
  & \min_{w} F(w) = \frac{1}{n} \sum_{k=1}^{n} \mathcal{L} (\hat{y}_k, y_k),              \nonumber
\end{IEEEeqnarray}
where $y_k$ denotes the label of the $k$-th graph node and $\mathcal{L}$ represents the loss function.
The above optimization process proceeds for multiple iterations until it satisfies the stopping criteria.

\subsection{Graph Federated Learning}
We focus on the node classification task in the context of GFL. As depicted in \figurename \ref{GFL:GFL}, we consider a GFL system constituted by $P$ clients and a central server orchestrating the overall process. Each client $i, \forall i\in [P]$ possesses its local data $\mathcal{D}_i = (G_i, Y_i) $, where $G_i = (\mathcal{V}_i, E_i, \mathbf{X}_i)$ is a subgraph of a global graph $G$, $n_{i}=|\mathcal{V}_{i}|$ represents the number of graph nodes, and $Y_{i} = \{ y_{i,k}, k\in[n_{i}] \}$ represents the labels of graph nodes. At each round, $K$ clients are selected by the central server to participate in the GFL process. Each client trains its local model on its own subgraph for $E$ iterations, aiming to minimize the local empirical risk function $F_i(w)$
\begin{IEEEeqnarray}{lc}\label{GFLtwo}
  & \min_{w} F_i(w) = \frac{1}{n_i} \sum_{k=1}^{n_i} \mathcal{L}_i (\hat{y}_{i,k}, y_{i,k}), \quad  i=1, \dots, K.              \nonumber
\end{IEEEeqnarray}

Then, the overall GFL task can be formulated as solving a distributed optimization problem as
\begin{IEEEeqnarray}{lc}\label{GFLRisk}
  & \min_{w} F(w) = \text{Agg}( \{ F_i(w), i \in [P] \}),              \nonumber
\end{IEEEeqnarray}
where $\text{Agg}(\cdot)$ is the aggregation function (e.g., the average function in FedAvg).

\subsection{Definitions and Assumptions}\label{subsec:defi}
As mentioned, this paper focuses on the overlapping subgraph data setting. Here, we formally define overlapping graph nodes and overlapping links, as well as overlapping ratios for both.

\begin{definition}[Overlapping Graph Nodes]
    If there are overlapping graph nodes between client $i$ and the other clients, i.e.
    \begin{IEEEeqnarray}{cl}
        \mathcal{V}_i \cap (\bigcup_{k\in [P], i \neq k} \mathcal{V}_k) \neq \varnothing.
    \end{IEEEeqnarray}
    the client $i$ can be termed to feature overlapping graph nodes.
    If there exists at least one client featuring overlapping graph nodes, the GFL settings are termed to feature overlapping graph nodes.
\end{definition}

\begin{definition}[Overlapping links]
    If the $i$-th client features overlapping graph nodes and there exist overlapping links among these graph nodes, then the $i$-th client can be termed to feature overlapping links.
    The presence of at least one client featuring overlapping links indicates the existence of overlapping links in the GFL settings.
\end{definition}

Based on the definition of overlapping data, we can formalize the graph node overlapping ratios and link overlapping ratios as follows.
\begin{definition}[Graph Node Overlapping Ratio]
    If there exist overlapping graph nodes between client $i$ and client $k$, the graph node overlapping ratio $\mathbf{N}_{i,k}$ of the client $i$ to client $k$ can be defined by the number of overlapping graph nodes divided by the total number of graph nodes on client $i$. Formally, it can be expressed as
    \begin{IEEEeqnarray}{cl}
        \mathbf{N}_{i,k} = \frac{|\mathcal{V}_{i} \cap \mathcal{V}_{k}|}{|\mathcal{V}_{i}|},
    \end{IEEEeqnarray}
    The average graph node overlapping ratios of client $i$ to all the other clients can be defined as
    \begin{IEEEeqnarray}{cl}
        \mathbf{N}_{i} = \frac{1}{P-1} \sum_{k \in [P], i\neq k} \mathbf{N}_{i,k}.
    \end{IEEEeqnarray}
    The average overlapping ratios in GFL can be defined as
    \begin{IEEEeqnarray}{cl}
      \mathbf{N} & = \frac{1}{P} \sum_{i=1}^{P} \mathbf{N}_i.
    \end{IEEEeqnarray}
\end{definition}

\begin{definition}[Link Overlapping Ratio]\label{def:link-over-rate}
    For each client $i\in[P]$, the average link overlapping ratio to all other clients is defined as
    \begin{IEEEeqnarray}{cl}
        \mathbf{T}_{i} = \frac{1}{P-1} \sum_{k \in [P], i\neq k} \mathbf{T}_{i,k},
    \end{IEEEeqnarray}
    where $\mathbf{T}_{i,k} = \frac{ | E_{i} \cap E_{k} | }{ |E_{i}| }$ represents the link overlapping ratio of the client $i$ to client $k$.
\end{definition}

These formalized ratios provide quantitative measures to assess the overlapping degree of subgraphs, offering insights into the shared node characteristics and connectivity among different clients.
Benefiting from the above measures, the settings with varying graph node and link overlapping ratios across clients can be featured as \textit{imbalanced} overlaps.
The clients with higher overlapping ratios commonly tend to achieve better model performance, as presented in Section \ref{sec:motivation}. We refer to these clients as advantaged clients. Conversely, clients with low or no overlapping data are referred to as disadvantaged clients, as they exhibit comparatively lower model performance.

For the convenience of the subsequent theoretical analysis, some assumptions on loss functions are given as follows.
\begin{assumption}[$L$-smoothness]\label{assume-Lsmooth}
 \textit{The differentiable objective function $F(w)$ is $L$-smooth:}
  $\forall w_1, w_2$, $\exists L \in \mathbb{R}$, s.t.
\begin{IEEEeqnarray}{cl}
  \langle \nabla F(w_1), w_1-w_2 \rangle \leq F(w_1) - F(w_2) + \frac{L}{2} ||w_1-w_2||^2.  \notag
\end{IEEEeqnarray}
\end{assumption}

\begin{assumption}[Locally unbiased gradients and bounded variance]\label{assume-local}
\textit{The gradient $\nabla f(w_{j,i}, \xi_{j,i})$ of client $i$ is client-level unbiased, that is,}
\begin{IEEEeqnarray}{cl}
  \mathbb{E} [\nabla f(w_{j,i}, \xi_{j,i}]  = \nabla F_i(w_j).
\end{IEEEeqnarray}
\textit{The gradient $\nabla F(w_{j,i}, \xi_{j,i})$ of client $i$ has client-level bounded variance: $ \exists$  $G_{l}^{2} \in \mathbb{R}, $}
\begin{IEEEeqnarray}{cl}
  \mathbb{E} [||\nabla f(w_{j,i}, \xi_{j,i}) - \nabla F_i(w_j)||^2]  \leq G_{l}^{2}.
\end{IEEEeqnarray}
To guarantee the convergence of the model, we also need to assume $\nabla F_i(w_j)$ satisfies global-level bounded variance: $\exists$ $G_{g}^{2} \in \mathbb{R}$,
\begin{IEEEeqnarray}{cl}
  \mathbb{E} [||\nabla F_i(w_j) - \nabla F(w_{j})||^2] \leq G_{g}^{2}.
\end{IEEEeqnarray}
\end{assumption}

\section{Motivation: Unfairness Caused by Imbalanced Overlaps}\label{sec:motivation}
Section~\ref{subsec:motivateExp} first provides some empirical observations on performance inconsistency across clients resulting from overlapping data. Subsequently, Section~\ref{subsec:motivateAnalysis} reasons about such unfairness issues from a theoretical perspective.

\subsection{Empirical Observations}\label{subsec:motivateExp}

\begin{figure}[!t]
  \centering
  \subfigure[Loss on training set.]{
		\includegraphics[width=1.5in]{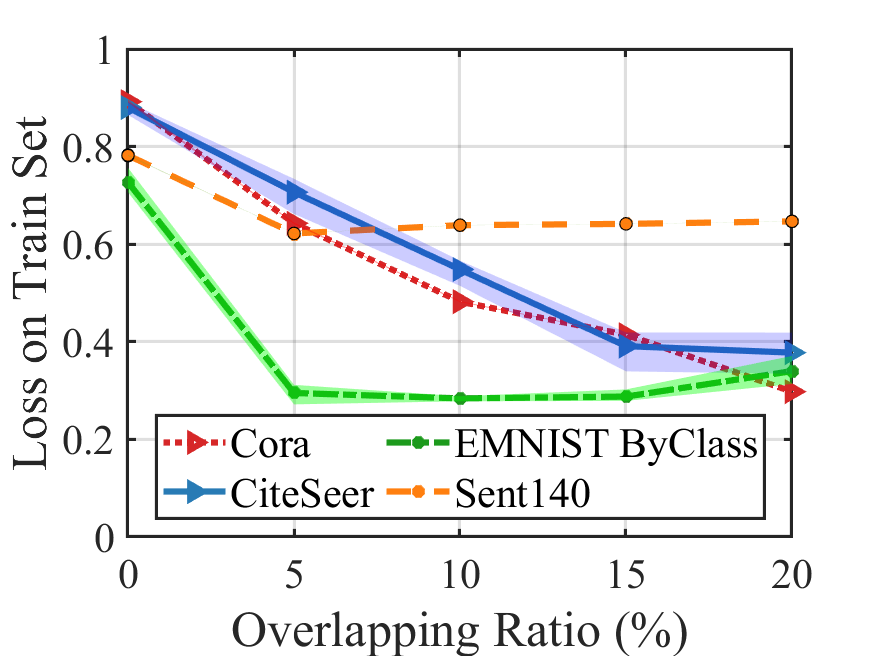}
		\label{moti:perf:a}
	}
  \hfil
  \subfigure[Loss on test set.]{
		\includegraphics[width=1.5in]{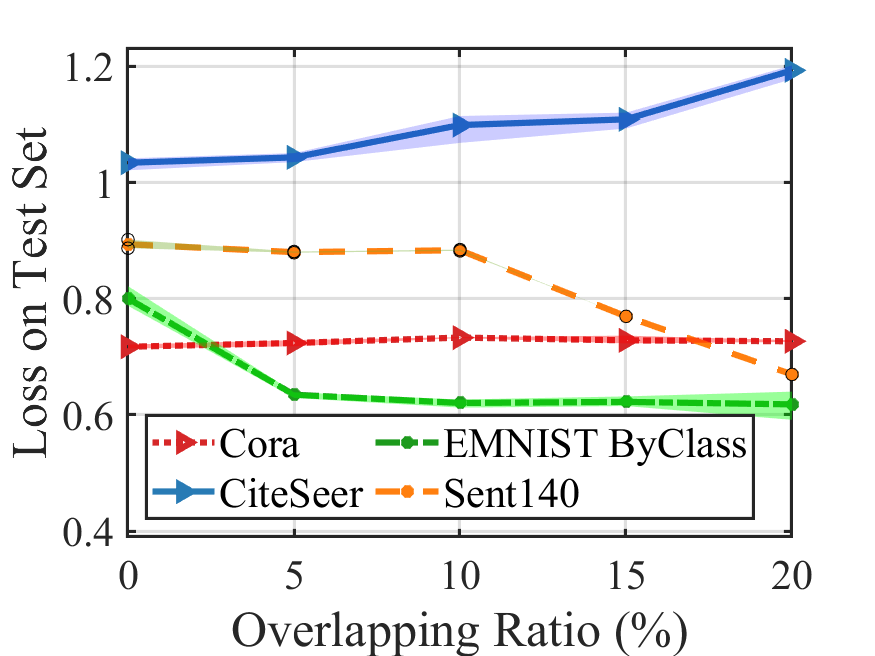}
		\label{moti:perf:b}
	}
   \caption{Model losses on overlapping graph data (Cora, CiteSeer) and non-graph datasets (EMNIST ByClass, Sent140).}\label{moti:perf}
\end{figure}

\begin{figure}[!t]
  \centering
  \subfigure[Variance of training loss.]{
		\includegraphics[width=1.5in]{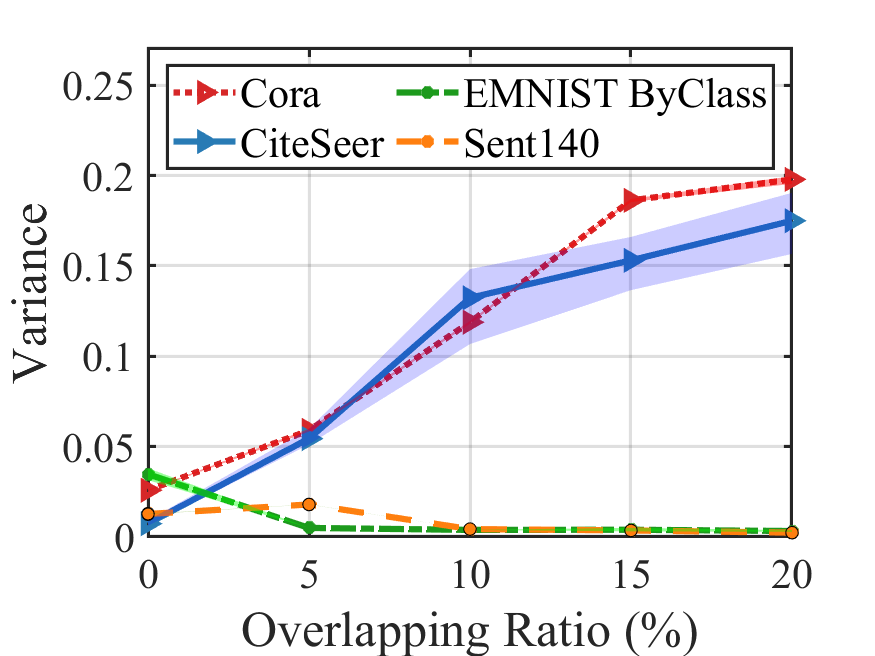}
		\label{moti:fair:a}
	}
  \hfil
  \subfigure[Entropy of training loss.]{
		\includegraphics[width=1.5in]{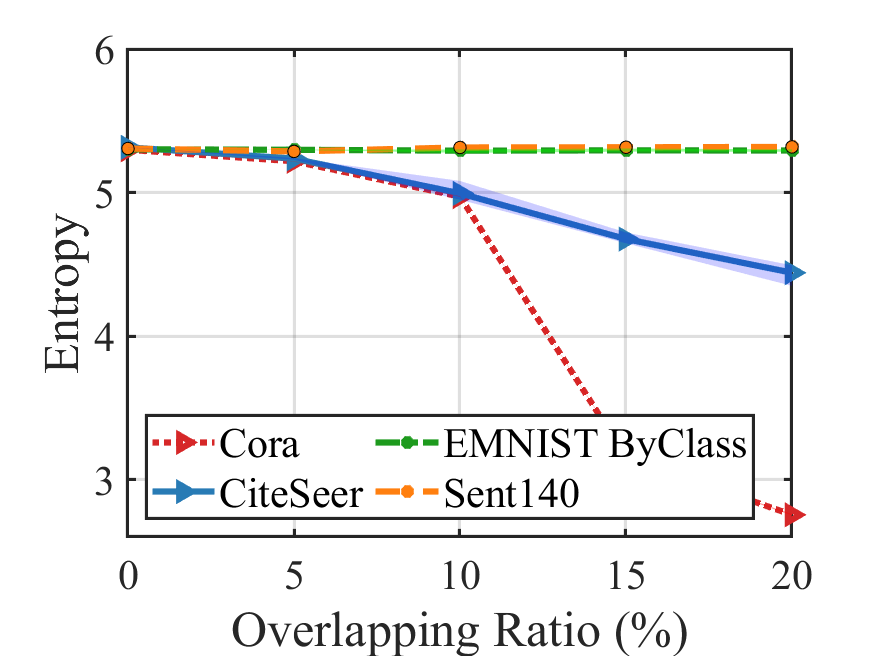}
		\label{moti:fair:b}
	}
 \caption{Performance inconsistency across clients on overlapping graph data (Cora, CiteSeer) and non-graph datasets (EMNIST ByClass, Sent140).}\label{moti:fair}
\end{figure}

To investigate the impact of imbalanced overlaps, we deployed FedAvg on both graph data (Cora, CiteSeer) and non-graph datasets (EMNIST ByClass, Sent140) using classical models, including GCN, CNN, and LSTM. We set the average overlapping ratios $\mathbf{N} \in \{ 0, 0.05, 0.10, 0.15, 0.20 \} $, where $\mathbf{N}=0$ represents the non-overlapping settings. The overlapping levels among clients are categorized into three groups: no overlaps (0), low overlaps ($\mathbf{N}$), and high overlaps ($2\mathbf{N}$).

For non-graph data (EMNIST ByClass and Sent140), as shown in \figurename~\ref{moti:perf},  the model losses exhibit a significant decrease on both the training set and test set as the average overlapping ratio increases. This indicates that the overlapping data brings model performance gains on both overlapping and non-overlapping data. \figurename~\ref{moti:fair} displays that the variance decreases obviously and the entropy remains relatively stable with increasing overlapping ratios. This is due to the fact that the model performance gains on overlapping and non-overlapping data are similar. Therefore, the imbalanced overlapping ratios across clients result in a slight impact on fairness.

For graph data (Cora and CiteSeer), \figurename~\ref{moti:perf} illustrates that the losses exhibit a significant decrease on the training set and an increase on the test set with the increasing average overlapping ratio.
\figurename~\ref{moti:fair} shows that the variance of losses tends to increase and the entropy decreases sharply with higher overlapping ratios. This indicates a serious unfairness issue across clients. The reasons are as follows
\begin{itemize}
    \item Reduced levels of heterogeneity among overlapping subgraphs: The diversity in node features and graph structures results in more complex heterogeneous scenarios compared to non-graph data. The high level of overlap among clients reduces the heterogeneity, effectively improving model performance on the training set and advantaged clients.

    \item More comprehensive edge information for overlapping nodes: The presence of missing edges among nodes in subgraphs leads to degraded model performance. However, the diverse edge features in multiple subgraphs for overlapping nodes to some extent compensate for this deficiency.
\end{itemize}
Therefore, the imbalanced overlaps result in the differentiated impact on the performance gains for overlapping and non-overlapping data. Clients with lower overlapping ratios derive little profit or even suffer from the decreasing model utility on unseen data.

In summary, there are two key observations:
\textit{First, imbalanced overlapping data among clients gives rise to unfairness concerns.
Second, the unfairness issue is particularly pronounced in the context of graph data but exhibits uncertainty for non-graph datasets.}

\subsection{Theoretical Reasoning}\label{subsec:motivateAnalysis}

To reason about the empirical observations, we further conduct a comprehensive theoretical analysis considering the impact of imbalanced overlaps. Specifically, we first present the common impact of imbalanced overlaps on the loss function to demonstrate the unfairness issues for both graph and non-graph data. Then, we analyze the differentiated impacts of imbalanced overlaps on the upper bound of fairness metric for non-graph and graph data separately. Finally, we conclude the analysis in Remark.

\textit{\textbf{For both graph and non-graph data, model overfitting on imbalanced overlapping data incurs a negative impact on fairness.}} To illustrate the impact of overlapping data, we begin by considering non-graph data. Assume that the client $i \in [P]$ possesses a collection of local data $\mathcal{D}_i = \{ d_{i,1}, \ldots, d_{i,n_i} \}$, where $n_i$ is the number of local data.
We formally define the overlaps among clients as follows
\begin{IEEEeqnarray}{cl}
   \exists \ i, k \in [P], s.t. \quad \mathcal{D}_{i} \cap \mathcal{D}_{k} \neq \emptyset.   
\end{IEEEeqnarray}
The overall data $\mathcal{D}$ contains both overlapping and non-overlapping data, which can be divided into two parts, i.e.
\begin{IEEEeqnarray}{cl}
  \mathcal{D} & = \mathcal{D}_o' \bigcup \mathcal{D}_n' = \{ d_{1}^{o}, d_{2}^{o}, \ldots, d_{n_1'}^{o} \} \bigcup \{ d_{1}^{n}, d_{2}^{n}, \ldots, d_{n_2'}^{n} \},   \nonumber
\end{IEEEeqnarray}
where $\mathcal{D}_o'$ is the overlapping data and corresponding overlapping frequency is $\{ o_{1}, o_{2}, \ldots, o_{n_1'} \}$, and $\mathcal{D}_n'$ is the non-overlapping data.
Then, the FL objective is
{\small
\begin{IEEEeqnarray}{cl}\label{equ:emp_loss}
   &  \min_{w} \frac{1}{n} \left( \sum_{d_{k_o}^o \in \mathcal{D}_o'} o_{k_o} F(w, d_{k_o}^o) + \sum_{d_{k_n}^n \in \mathcal{D}_n'} F(w, d_{k_n}^n) \right).
\end{IEEEeqnarray}
}

\textit{\textbf{For non-graph data, the overlapping data can mitigate data heterogeneity and incur a positive impact on fairness.}} Assume that the loss function $F(w, d)$ satisfies $L_{d}$-Lipschitz continuity with respect to local data, i.e. $\forall \ d_{i}, d_{j} \in \mathcal{D}$, s.t.
\begin{IEEEeqnarray}{cl}\label{moti-eq:Lcontinuity}
    ||F(w,d_{i}) - F(w,d_{j})|| & \leq L_{d} ||d_{i} - d_{j}||.
\end{IEEEeqnarray}
\textbf{- Similar model performance gain for both overlapping and non-overlapping data.} As shown, the loss disparity is determined by the difference between data distributions.
The loss disparity between overlapping data $d_{i}^{o}$ and non-overlapping data $d_{j}^{n}$ with the same labels can be bounded as:
\begin{IEEEeqnarray}{cl}\label{moti-eq:similardata}
    \| F(w, d_{i}^{o}) - F(w, d_{j}^{n}) \| & \leq L_{d} \|d_{i}^{o} - d_{j}^{n} \|.
\end{IEEEeqnarray}
In cases where data sharing the same labels also share the same distribution, $\|d_{i}^{o} - d_{j}^{n} \|$ is a small term. Since the model overfits the overlapping data $d_{i}^{o}$, the loss $F(w, d_{i}^{o})$ on $d_{i}^{o}$ should be small. Consequently, according to Eq.~\eqref{moti-eq:similardata}, the loss $F(w, d_{j}^{n})$ on non-overlapping data $d_{i}^{n}$ is also controlled to be a small value. Therefore, overfitting of overlapping data improves the model utility on non-overlapping data (as illustrated in \figurename~\ref{moti:perf}) reducing performance inconsistency and mitigating unfairness (as illustrated in \figurename~\ref{moti:fair}).

\noindent\textbf{- Reduced levels of heterogeneity among clients.} To further observe the impact of overlaps on fairness,  we also give a proxy of the fairness,  the average Euclidean distance between the loss of each local model, and the average loss of all local models:
{\small
\begin{IEEEeqnarray}{cl}\label{moti-eq:fair}
    & \frac{1}{P} \sum_{i=1}^{P} || F(w, \mathcal{D}_{i}) - \frac{1}{P} \sum_{k=1}^{P} F(w, \mathcal{D}_{k}) ||,
\end{IEEEeqnarray}
}

\noindent then we explore the impact of overlaps on the distance. Based on Eq.~\eqref{moti-eq:Lcontinuity}, the distance can be bounded as follows: 
{\small
\begin{IEEEeqnarray}{cl}\label{moti-eq:comp}
    & \frac{ L_{d} }{n^2} \left(\sum_{d_{k_o^1}^{o}, d_{k_o^2}^{o} \in \mathcal{D}_{o}'}  \|  d_{k_o^1}^{o} - d_{k_o^2}^{o} \| +  \sum_{d_{k_n^1}^{n}, d_{k_n^2}^{n} \in \mathcal{D}_{n}'}  \|  d_{k_n^1}^{n} - d_{k_n^2}^{n} \| \right.   \nonumber  \\
    & \quad  \left.  + \sum_{d_{k_o}^{o} \in \mathcal{D}_{o}'} \sum_{d_{k_n}^{n} \in \mathcal{D}_{n}'}  \|  d_{k_o}^{o} - d_{k_n}^{n} \|   \right).
\end{IEEEeqnarray}
}

\noindent The derivation procedure is referenced in Appendix~\ref{sec:moti}. Eq~\eqref{moti-eq:comp} indicates that performance inconsistency is primarily determined by data disparity. In overlapping settings, the data disparity among non-overlapping data remains unaltered, while the data disparity among overlapping data is significantly narrowed. As a consequence, overlapping data reduces the upper bound of this fairness proxy.

\textit{\textbf{For graph data, the overlapping subgraphs cannot alleviate data heterogeneity effectively.}}
Assume that the loss function $F(w, d)$ for graph data satisfies $\forall \ d_{i}, d_{j} \in \mathcal{D}$, s.t.
{\small
\begin{IEEEeqnarray}{l}\label{moti-eq:graph-Lcontinuity}
    ||F(w,d_{i}) - F(w,d_{j})||_2 \leq L_{1}^{g} ||d_{i} - d_{j}||_2 + L_{2}^{g}(\mathcal{N}_{i}, \mathcal{N}_{j}),   \nonumber
\end{IEEEeqnarray}
}
\noindent where $\mathcal{N}_{i}$ represents the neighboring graph nodes of the $i$-th graph node,  $L_{2}^{g}(\cdot, \cdot)$ is a function of the neighboring graph nodes to bound the loss disparity resulting from links.
{\\ \textbf{- More comprehensive edge information and model performance gain for overlapping nodes.}} We can observe that the overfitting only improves the model utility for graph nodes with similar neighboring graph nodes. However, the diverse links may result in unpredictable heterogeneity and bring little model performance gain for non-overlapping nodes.
{\\ \textbf{- Reduced levels of heterogeneity only for clients with higher overlapping ratios.}} The average Euclidean distance between the loss of each local model and the average loss of all local models is bounded as follows:
{\small
\begin{IEEEeqnarray}{cl}\label{moti-eq:graph-comp}
    & \frac{1}{n^2} \left( \sum_{d_{k_o^1}^{o}, d_{k_o^2}^{o} \in \mathcal{D}_{o}'} \left( L_{1}^{g} \|  d_{k_o^1}^{o} - d_{k_o^2}^{o} \| +  L_{2}^{g} (\mathcal{N}_{d_{k_o^1}^{o}}, \mathcal{N}_{d_{k_o^2}^{o}}) \right) \right. \nonumber  \\
    & \left.  + \sum_{d_{k_o}^{o} \in \mathcal{D}_{o}'} \sum_{d_{k_n}^{n} \in \mathcal{D}_{n}'}  \left( L_{1}^{g}  \|  d_{k_o}^{o} - d_{k_n}^{n} \|   +  L_{2}^{g} (\mathcal{N}_{d_{k_o}^{o}}, \mathcal{N}_{d_{k_n}^{n}}) \right) \right.   \nonumber  \\
    & \left. + \sum_{d_{k_n^1}^{n}, d_{k_n^2}^{n} \in \mathcal{D}_{n}'}  \left( L_{1}^{g}  \|  d_{k_n^1}^{n} - d_{k_n^2}^{n} \| +  L_{2}^{g} (\mathcal{N}_{d_{k_n^1}^{o}}, \mathcal{N}_{d_{k_n^2}^{o}}) \right) \right).
\end{IEEEeqnarray}
}

\noindent Although overlaps diminish the average distance among graph nodes, the uncertainty of the term $L_{2}^{g}(\cdot, \cdot)$ associated with adjacent graph nodes remains substantial. \textit{The trained model prefers clients with higher overlapping ratios or specific distributions} (e.g. the distributions of graph nodes resembling the overlapping graph nodes and sharing similar distribution of neighboring graph nodes.), \textit{thereby leading to serious unfairness}. The variations in node features and adjacency matrices result in various levels of data heterogeneity. Therefore, the upper bound of this fairness proxy becomes highly uncertain.

\vspace{0.18cm}

\noindent \textbf{Remark:}
{\textit{Both the empirical observations and theoretical analysis demonstrate two insights: }}
\begin{enumerate}[leftmargin = *]
  \item \textit{Unfairness issue arises from imbalanced overlaps.} {\rm{Imbalanced overlaps result in differentiated weights across clients, further leading to unfairness. }}
  \item \textit{Unfairness issue is more pronounced in GFL.} {\rm{This is attributed to the intricate impact on data heterogeneity introduced by diverse graph nodes and links, which would not be encountered in non-graph datasets.}}
\end{enumerate}

\section{FairGFL Algorithm}\label{sec:method}

We have identified the unfairness issue caused by imbalanced overlapping subgraphs across clients in Section~\ref{sec:motivation}. To address it, we propose a novel \underline{FAIR}ness-aware \underline{G}raph \underline{F}ederated \underline{L}earning (FairGFL) algorithm to mitigate such unfairness while ensuring model utility.

\subsection{Main Idea}\label{subsec:main-idea}

As discussed in Section~\ref{subsec:motivateAnalysis}, the unfairness issue arises from the imbalanced weights assigned to different local models' losses due to the imbalanced overlapping ratios. To address this, we propose a fairness-aware aggregation strategy considering the imbalanced overlapping ratios across clients. Specifically, we reweight different local models' losses by assigning weights negatively related to the overlapping ratios.
However, it still remains a challenge to estimate overlapping ratios for the server without access to the raw data~\cite{triastcyn2019federated, seif2020wireless}. To overcome this challenge, we propose a privacy-preserving method that enables the server to accurately estimate the overlapping ratios using samples sanitized through an LDP protocol.

Despite the enhanced fairness, the fairness-aware aggregation may compromise the global model utility. Specifically, while disadvantaged clients experience minor improvements, other advantaged clients may suffer a significant decline in model utility. To address this issue, we design a loss-related regularizer to weaken the role of fairness and prevent a substantial decline in model utility for advantaged clients. In this way, FairGFL improves the tradeoff between fairness and model utility.

\subsection{FairGFL Framework}\label{subsec:system-model}
\begin{figure*}[!t]	
  \centering	
  \includegraphics[width=0.85\textwidth, trim={29 130 30 154}, clip]{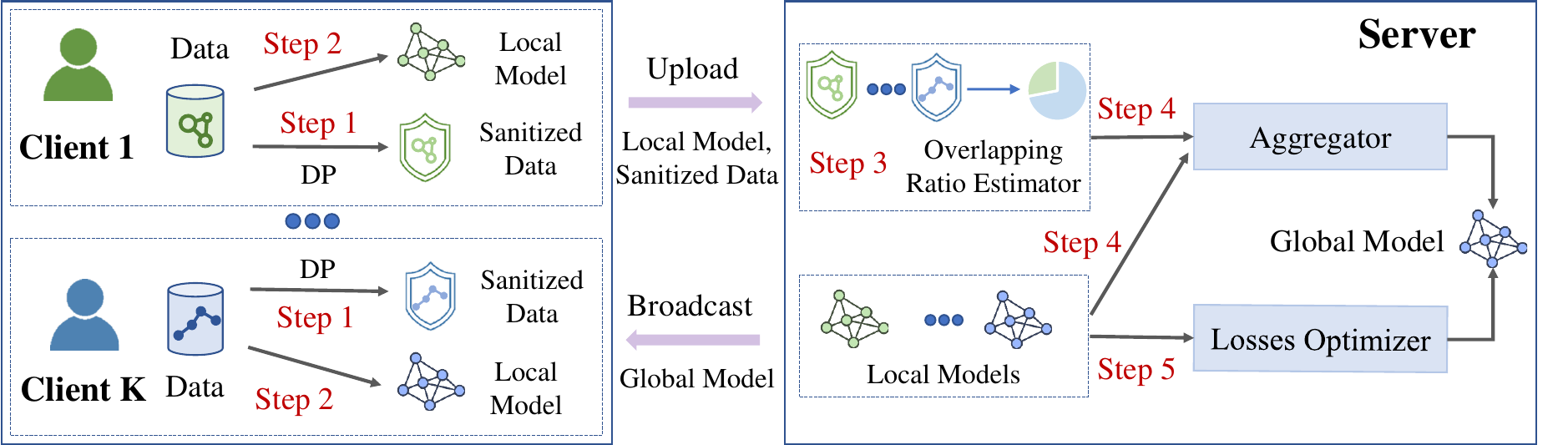}   
  \caption{The workflow of FairGFL. FairGFL comprises five main steps: privacy-preserving data perturbation (Step 1), local training (Step 2), overlapping ratio estimation (Step 3), fairness-aware aggregation (Step 4), and composite losses optimization (Step 5).}
  \label{SystemModel}		
\end{figure*}

\begin{algorithm}[!t]
  \caption{FairGFL: Client Side.}\label{Clients}
  \begin{algorithmic}[1]
    \REQUIRE Mini-batch size $b$, local iterations $E$.
    \ENSURE Privacy-preserving mini-batch subgraph nodes $\mathcal{V}_{i,j}$, randomized mini-batch adjacent matrix $\mathbf{E}_{i,j}$, local model $w_{j,i}^{E}$.
    \STATE Receive $w_j$ from the server and initialize $w_{j,i}^{0}=w_{j}$.
    \FOR{$e = 0, 1, \ldots, E-1$}
      \STATE $w_{j,i}^{e+1} = w_{j,i}^{e} - \eta f(w_{j,i}^{e}, \xi_{i,j})$.
    \ENDFOR
    \STATE $w_{j,i} \gets w_{j,i}^{E}$.
    \STATE Sanitize the mini-batch reduced-dimensional subgraph nodes and adjacent matrix based on Eq.~\eqref{eq:nodenoise} and Eq.~\eqref{eq:link-noise}.
    \STATE Sparsify the noised adjacent matrix.
    \STATE Upload $\mathcal{V}_{i,j}$, $\mathbf{E}_{i,j}$ and $w_{j,i}$.
  \end{algorithmic}
\end{algorithm}

\begin{algorithm}[!t]
	\caption{FairGFL: Server Side.}\label{Server}
	\begin{algorithmic}[1]
		\REQUIRE Initialized model $w_{0}$, number of sampled clients $K$ at each round, coupling weight $\alpha$, accumulated weight $\beta$.
		\ENSURE Optimal global model $w^*$.
		\FOR {$j=1,2,\ldots$}
		\STATE Broadcast $w_{j}$ to sampled $K$ clients $\mathcal{S}_{j}^{c}$.
        \STATE Receive $\{\mathcal{V}_{i,j}$, $\mathbf{E}_{i,j}$, $w_{j,i}\}$ from each client $i \in \mathcal{S}_{j}^{c}$.
		\STATE Compute the graph node and link overlapping ratios $\mathbf{N}$ and $\mathbf{T}$ according to Theorem \ref{TH:node-over-rates} and Theorem \ref{TH:link-over-rates}.
        \STATE $\mathbf{N}_{j}^{a} = \beta \mathbf{N}_{j} + (1-\beta) \mathbf{N}_{j}^{a}$.
        \STATE $\mathbf{T}_{j}^{a} = \beta \mathbf{T}_{j} + (1-\beta) \mathbf{T}_{j}^{a}$.
        \STATE $\mathbf{O}_{j} = \alpha \mathbf{N}_{j}^{a} + (1-\alpha) \mathbf{T}_{j}^{a}$.
		\STATE $w_{j} =  \frac{1}{K} \sum_{i=1}^{K} \frac{1}{1+\mathbf{O}_{i}} w_{i,j}$.
		\STATE Update the global model $w_{j}$ for one step by optimizing the loss function in Eq.~\eqref{eq:loss-all}.
        \STATE $j \gets j+1$.
        \ENDFOR
	\end{algorithmic}
\end{algorithm}

FairGFL framework consists of two main components, including fairness-aware aggregation and composite losses optimization. To achieve cross-client fairness, the server requires access to the clients' overlapping ratios. However, this could compromise privacy, thereby violating the principle of FL. To safeguard data privacy, we propose privacy-preserving data perturbation and overlapping ratio estimation methods. The workflow of FairGFL is illustrated in \figurename~\ref{SystemModel}. The pseudo-code of FairGFL consists of two parts, Algorithm~\ref{Clients} for clients and Algorithm~\ref{Server} for the server.

For clients (Steps 1-2), as described in Algorithm \ref{Clients}, they first download the current model parameters from the server to be their local models (Line 1). Then, they update local models for $E$ iterations (Line 2) and upload these models to the server (Line 5). To estimate the overlapping ratios and achieve fairness, clients also upload privacy-preserving mini-batch subgraphs (Lines 3-5).

For the server (Steps 3-5), as shown in Algorithm \ref{Server}, after initializing the model (Line 1), the server broadcasts it to a set of uniformly sampled $K$ clients $\mathcal{S}_{j}^{c}$ for local computation (Lines 3-4). Once the $K$ clients complete local training and upload their models, the server estimates the overlapping ratios (Lines 5-9) and aggregates the $K$ local models based on the overlapping ratios (Line 10). Furthermore, the global model is updated for one step by optimizing the composite losses function to improve the model utility (Line 11).

In Sections \ref{Sec:method-sec1} to \ref{Sec:method-sec4}, we present detailed descriptions and rationale of four novel components (Step 1 and Steps 3-5) of FairGFL.

\subsubsection{Privacy-preserving data perturbation}\label{Sec:method-sec1}

To mitigate the impact of imbalanced overlaps, one intuitive method is to remove redundant overlapping subgraphs (graph nodes and links) but one. However, this method faces two challenges.
\emph{On one hand}, the server deprives itself of detecting and filtering out the overlapping subgraphs due to no access to the raw data.
\emph{On the other hand}, some non-overlapping link information will be lost upon the removal of the overlapping subgraphs, since the local subgraphs may only partially overlap in terms of graph nodes.

To address the above two challenges, each client can upload a subgraph by sampling a mini-batch of graph nodes and their corresponding links. Nevertheless, clients are reluctant to upload the raw data in consideration of privacy disclosure. To relieve their concerns, we propose a novel LDP technique~\cite{ren2022ldp} to enhance the privacy of local subgraphs. The rules for sanitizing the graph nodes and links are as follows.

\noindent\emph{\textbf{Encode and obfuscate the graph nodes.}}
Each client first utilizes an encoder~\cite{yu2021auto} to process the graph node feature vectors. Prior to training the FL model, the server broadcasts an encoder to all the clients and each client encodes their local mini-batch graph nodes. The encoder exhibits two advantages.
\emph{On one hand}, it reduces the communication cost by decreasing the dimension of the raw features. The graph node feature vectors are typically high-dimensional and the transmission of the entire features consumes overwhelming communication resources.
\emph{On the other hand}, the encoder extracts important information while making it difficult to perfectly recover the raw features. However, the honest but curious server could potentially leverage a well-designed decoder to recover private information~\cite{cao2022perfed, gui2021review}.

To provide strict privacy protection, we propose a novel LDP technique to sanitize encoded graph node features, enabling accurate estimation of overlapping ratios.
Previous research~\cite{sun2019analyzing} applies the LDP technique to indices of graph nodes to obfuscate the membership information of graph nodes. However, the server needs to assign indices to all the graph nodes before training, which may result in privacy leakage. Besides, another research~\cite{lin2022towards} proposes to randomize some elements in high-dimensional graph node features to values of -1 or 1 and map other elements to 0, which is too coarse to achieve accurate estimation. Differently, we propose a more fine-grained mechanism, denoted as $\mathcal{M}_{N}$. In $\mathcal{M}_{N}$, each client independently randomizes each element of encoded features to a term in the equidistant $p$ quantile sequence according to the distance from the minimum and maximum values of the encoded features. Specifically, each element can be perturbed as $ \frac{i}{p}, i\in \{0,1,\ldots,P\} $ with a probability of
\begin{IEEEeqnarray}{cl}\label{eq:nodenoise}
  \frac{e^{\epsilon_a/p}-1}{e^{(p+1)\epsilon_a/p}-1} \cdot & e^{\epsilon_a ( 1 - \frac{1}{p}\lfloor  p |\frac{x_{i,j}-x_{\min}}{x_{\max} - x_{\min}} - \frac{i}{p} | \rfloor ) },
\end{IEEEeqnarray}
where $\epsilon_a$ represents the privacy budget for each protected element $x_{i,j}$ of the reduced-dimensional features, $p$ is the preset sophisticated level, and $[x_{min}, x_{max}]$ defines the domain range of the feature.

\begin{theorem}\label{TH:nodenoise}
  The mechanism $\mathcal{M}_{N}$ to obfuscate the graph nodes satisfies $\epsilon_a$-LDP.
\end{theorem}
\begin{IEEEproof}\label{proof:nodenoise}
  Please refer to Appendix \ref{ap:proof:nodenoise}.
\end{IEEEproof}
\noindent\emph{\textbf{Perturb the links.}}
In a graph, links are typically represented as an adjacent matrix, where each element is a binary value. We denote the adjacent matrix for client $i$ as $\mathbf{E}_i = (e_{p,q}^{i})$, $p, q \in [n_i]$, where $e_{p,q}^i = 1$ indicates the presence of a link between the $p$-th graph node and $q$-th graph node. To perturb the links, we adopt a mechanism $\mathcal{M}_{E}$ \cite{yang2020secure, qin2017generating, wei2020asgldp}. Specifically, given a privacy budget $\epsilon_b$, each user flips each element in the adjacent matrix with a probability $p_{e}$, resulting in a randomized adjacent matrix $\tilde{\mathbf{E}}_i = (\tilde{e}_{p,q}^{i})$. The flipping rule for each element $e_{p,q}^{i}$ is as follows.
\begin{IEEEeqnarray}{rl} \label{eq:link-noise}
  \tilde{e}_{p,q}^{i} =
  \begin{cases}
    e_{p,q}^{i},  & with\quad prob.\quad 1-p_{e}=\frac{e^{\epsilon_b}}{1+e^{\epsilon_b}}  \\
    1 - e_{p,q}^{i},  & with\quad prob.\quad p_{e} = \frac{1}{1+e^{\epsilon_b}}
  \end{cases}.
\end{IEEEeqnarray}

\begin{theorem}\label{TH:linknoise}
  The mechanism $\mathcal{M}_{E}$ to perturb the links satisfies $\epsilon_b$-LDP.
\end{theorem}
\begin{IEEEproof}\label{proof:linknoise}
  The proof methodology can refer to~\cite{lin2022towards}.
\end{IEEEproof}

Despite the provable privacy guarantee, the above flipping process would make the perturbed adjacent matrix too dense to accurately estimate overlapping ratios. Actually, some previous work~\cite{lin2022towards} has aimed to induce sparsity in the dense matrix, such as graph structure denoising, which, however, overlooks the graph characteristics and introduces extra noises.
To explore graph characteristics, we conducted statistical experiments on Cora and CiteSeer datasets to observe the relationship between the feature distance and the link existence of arbitrary two graph nodes.
Table~\ref{Tab:linknoise} presents the probability distribution of feature distance between two linked graph nodes. We can observe a sharply decreasing probability of link existence between graph nodes as their feature distance increases.

Inspired by the observation, we propose a correction approach to induce sparsity in the perturbed adjacent matrix.
Specifically, after perturbing the adjacent matrix, the client calculates the Euclidean distance between each pair of obfuscated graph nodes and estimates the proportion of 1s in the raw adjacent matrix according to Theorem~\ref{TH:link-correct}. That is, when the proportion of 1s in the perturbed adjacent matrix is $P_0$, the proportion of 1s in the raw adjacent matrix is estimated to be $ \frac{P_0 - p_{e}}{1-2p_{e}}$. Consequently, approximately $(P_0 - \frac{P_0 - p_{e}}{1-2p_{e}})$ proportional of 1s in the perturbed matrix will be corrected to 0s.
This correction approach effectively makes the adjacent matrix sparse. It is worth noting that the correction process is conducted on the perturbed graph node features and links, which would not incur extra privacy leakage due to the post-process property of DP.

\begin{table}[!ht]
  \centering
  \caption{Probability distribution of feature distance between linked graph nodes.}\label{Tab:linknoise}
  \begin{tabular}{l|cccccc}
    \toprule[0.5pt]
    \diagbox[width=8em,trim=l]{Datasets}{Distance}  &  0-10   &   11-20   &   21-30 &   31-40 & 41-50 & 51-60    \\   \midrule[0.1pt]
	Cora      &     \textbf{0.37}  &    0.27   &       0.17       &     0.12 & 0.05 & 0.02       \\
	CiteSeer  &        0.17  &  \textbf{0.66}   &     0.14         &   0.02  & 0.01 & 0.00    \\
    \bottomrule[0.5pt]
  \end{tabular}
\end{table}

\begin{theorem}\label{TH:link-correct}
  If a 0-1 matrix contains $100 \times x$ percent of 1s, and the flipping probability is set to $p_{e}$, then the expected proportion of 1s in the randomized matrix is given by $x+p_{e}-2x p_{e}$.
\end{theorem}
\begin{IEEEproof}
    By combining the definition of expectation and perturbation rule shown in Eq~\eqref{eq:link-noise}, we can easily deduce this conclusion.
\end{IEEEproof}

\subsubsection{Overlapping ratio estimation}\label{Sec:method-sec2}

To estimate the overlapping ratios, the server utilizes the sanitized graph node features and adjacent matrices to reconstruct the local subgraphs.

For graph node overlapping ratios, the server calculates the
Euclidean distance $||x_{i, i'} - x_{k,k'}||_1$ among noisy graph node features from different clients. If the distance is smaller than a predefined threshold, the two graph nodes are considered to represent the same data. In this way, the server estimates the overlapping ratio on mini-batch subgraphs of client $i$ to client $k$, denoted as $\tilde{\mathbf{N}}_{i,k}$.
Furthermore, according to Theorem~\ref{TH:node-over-rates}, the server can compute a $P\times P$-dimensional overlapping ratio matrix $\mathbf{N}$, where each element $\mathbf{N}_{i,k} = \frac{\tilde{\mathbf{N}}_{i,k} n_i}{b_k}$ denotes the graph node overlapping ratio of client $i$ to client $k$. However, in extreme scenarios where a part of clients sample most of the overlapping nodes into the minibatch while the other clients have massive graph nodes, the estimated overlapping ratios in $\mathbf{N}$ may exceed 1. In such cases, those values larger than 1 are set as 1 manually, then the corrected matrix is denoted as $\mathbf{N}^c$.

\begin{theorem} \label{TH:node-over-rates}
    Consider two clients $i$ and $k$, each with $n_i$ and $n_k$ graph nodes respectively. If clients $i$ and $k$ randomly sample $b_i$ ($b_{i} \leq n_{i}$) and $b_k$ ($b_{k} \leq n_{k}$) graph nodes, and the $b_i$ samples of client $i$ has a proportion $\tilde{\mathbf{N}}_{i,k} \in [0, 1]$ of samples overlapping with the $b_k$ samples of client $k$, then client $i$ has an expected proportion $\mathbf{N}_{i,k}$ of graph nodes overlapping with client $k$, where
    \begin{IEEEeqnarray}{c}
      \mathbf{N}_{i,k} =  \frac{\tilde{\mathbf{N}}_{i,k} n_i}{b_k} .
    \end{IEEEeqnarray}
\end{theorem}
\begin{IEEEproof}
    Please refer to Appendix~\ref{sec:appORE}.
\end{IEEEproof}

For link overlapping ratios, the server examines the overlapping links among overlapping graph nodes and computes the link overlapping ratio according to Definition~\ref{def:link-over-rate}. The link overlapping ratio matrix is denoted as $\tilde{\mathbf{T}}$, where each element $\tilde{\mathbf{T}}_{i,k}$ denotes the link overlapping ratio of client $i$ to client $k$. To estimate the link overlapping ratios among clients' subgraphs, we make one rational assumption. That is the overlapping graph nodes uploaded in each round are uniformly sampled from the overall overlapping graph nodes and the overlapping links are uniformly distributed among graph nodes. Then, according to Theorem~\ref{TH:link-over-rates}, the server can compute the overall link overlapping ratios matrix, defined as $\mathbf{T} = (\mathbf{T}_{i,k})$.

\begin{theorem}\label{TH:link-over-rates}
  Consider two clients, denoted as client $i$ and client $k$, each with $n_i$ and $n_k$ graph nodes respectively. Clients $i$ and $k$ select $b_i$ and $b_k$ samples randomly, and the $b_i$ samples of client $i$ have a proportion $\tilde{\mathbf{N}}_{i,k} \in [0, 1]$ of samples overlapping with those of client $k$. If the mini-batch link overlapping ratio of client $i$ to client $k$ is $\tilde{\mathbf{T}}_{i,k}$, the overall link overlapping ratio of client $i$ to client $k$ is $\frac{ \tilde{\mathbf{T}}_{i,k} n_{k}^{2} }{ b_i^2 }$.
\end{theorem}
\begin{IEEEproof}
    Please refer to Appendix~\ref{sec:appORE}.
\end{IEEEproof}

\subsubsection{Fairness-aware aggregation}\label{Sec:method-sec3}

Based on the estimated graph node overlapping ratios $\mathbf{N} = (\mathbf{N}_{i,k})$ and the link overlapping ratios $\mathbf{T} = (\mathbf{T}_{i,k})$, the server aggregates the model updates using weights negatively correlated with the estimated overlapping ratios. This aggregation strategy helps mitigate the impact of overlapping data.
However, accurately estimating the overlapping ratios is challenging due to the sampling randomness. From a statistical perspective, more accurate estimation can be achieved with a sufficient number of samples. Therefore, to avoid extreme cases, the estimated overlapping ratios in the current round are accumulated with the previous estimated overlapping ratios. The accumulated graph node and link overlapping ratios matrices in the $j$-th round are denoted as $\mathbf{N}_{j}^{a}$ and $\mathbf{T}_{j}^{a}$ respectively. Accordingly, the overall overlapping ratio $\mathbf{O}_{j} = \left( \mathbf{O}_{i,k}^{j} \right)$ can be calculated as follows
\begin{IEEEeqnarray}{c}  \label{eq:overall-over-rate}
  \mathbf{O}_{j} = \alpha \mathbf{N}_{j}^{a} + (1-\alpha) \mathbf{T}_{j}^{a},
\end{IEEEeqnarray}
where $\alpha \in [0, 1]$ is the coupling weight.

To enhance fairness, the server utilizes the adjusted weights\footnote{This work focuses on unfairness from overlapping data, assuming homogeneous data and client quality. For scenarios with high-quality clients or data, the proposed algorithm can be integrated with existing designed algorithms.} inversely proportional to the overlapping ratios to aggregate the received local models inspired by the Theorem~\ref{TH:final-weight} shown in Appendix~\ref{sec:appFA}.
Formally, the aggregating rule in the $j$-th round is
\begin{IEEEeqnarray}{rl} \label{eq:fairloss}
  w_{j} = & \frac{1}{K} \sum_{i=1}^{K} \frac{1}{1+\mathbf{O}_{i}} w_{i,j},
\end{IEEEeqnarray}
where $\mathbf{O}_{i} = \sum_{k \in [P], k \neq i} \mathbf{O}_{i,k}$ represents the overall overlapping ratios of client $i$.

\subsubsection{Composite losses optimization}\label{Sec:method-sec4}

The fairness-aware aggregation strategy improves cross-client fairness by reducing the model weights of clients with a significant number of overlapping subgraphs. However, this approach may lead to a performance loss for advantaged clients, resulting in a decrease in overall model utility. To maintain the model utility, we design a loss-related regularizer to control the worst model utility among clients.

Before presenting the ultimate optimization objective considering both fairness and model utility, we give an empirical composite losses function indicated by the fairness-aware aggregation shown in Eq.~\eqref{eq:fairloss} as follows.
\begin{IEEEeqnarray}{rl} \label{eq:fairlossRe}
  F(w) & = \frac{1}{K} \sum_{i=1}^{K} \frac{1}{1 + \mathbf{O}_{i}} F_i(w).
\end{IEEEeqnarray}
With the consideration of both fairness and the model utility, we introduce the loss-related regularizer $\lambda||\overrightarrow{\mathbf{F}}(w)||_{\infty}$ and obtain the ultimate loss function as follows.
\begin{IEEEeqnarray}{rcl} \label{eq:loss-all}
  F(w) & = & \underbrace{\frac{1}{K} \sum_{i=1}^{K} \frac{1}{1 + \mathbf{O}_{i}} F_i(w)}_{\text{I: Fairness}} + \underbrace{\lambda ||\overrightarrow{\mathbf{F}}(w)||_{\infty}}_{\text{II: Model Utility}},
\end{IEEEeqnarray}
where $\overrightarrow{\mathbf{F}}(w) = [F_1(w), F_2(w), \ldots, F_K(w)]^{\top}$ is a vector with each $j$-th element be the loss of client $j$. The regularization term plays a role in restraining fairness by a hyperparameter $\lambda$ and improves the model utility by minimizing the maximal losses. Therefore, in Eq.~\eqref{eq:loss-all}, the term I aims to improve fairness, while term II enhances model utility for all clients.

\section{Analysis}\label{sec:analysis}

In this section, we provide the communication and computation complexity and convergence analysis of the proposed FairGFL.

We first analyze the communication and computation complexity. In each round, each client $i$ transmits its local encoded nodes of dimension $d_1$ and the corresponding links based on its mini-batch of size $b_i$. Due to the sparsity of neighborhood connections, the communication complexity is $\mathcal{O}(d_1 b_i) + \mathcal{O}(b_i^2)$. The computational overhead on the client side primarily involves encoding and the LDP mechanisms on nodes and links. However, given the sparsity of links and the limited number of overlapping nodes, the additional computation for noising links is negligible. Therefore, the local computational complexity is dominated by $\mathcal{O}(b_i)$.

First, we define $i_{m}^{K} \triangleq \arg\max_{i \in [K]} \{ F_i(w) \}$, $i_{m}^{P} \triangleq \arg\max_{i \in [P]} \{ F_i(w) \}$, and $q_{j,i} = \frac{1}{1 + \mathbf{O}_{j,i}}$. In the $j$-th round, the expectation of the weighted losses can be expressed as
\begin{IEEEeqnarray}{cl}
   \frac{1}{K} \sum_{i=1}^{K} q_{j,i} F_i (w_{j}) + \lambda F_{i_m^K}(w_{j}).
\end{IEEEeqnarray}
The global composite empirical objective is
\begin{IEEEeqnarray}{cl}
    \min_{w} F(w) = \frac{1}{P} \sum_{i=1}^{P} q_{i} F_i (w) + \lambda F_{i_m^P}(w),
\end{IEEEeqnarray}
where $q_{i} = \frac{1}{ 1+\mathbf{O}_{i}^{T} }$.

\begin{figure*}[!t]
  \centering
  \includegraphics[scale=0.83, trim={55 67 56 68}, clip]{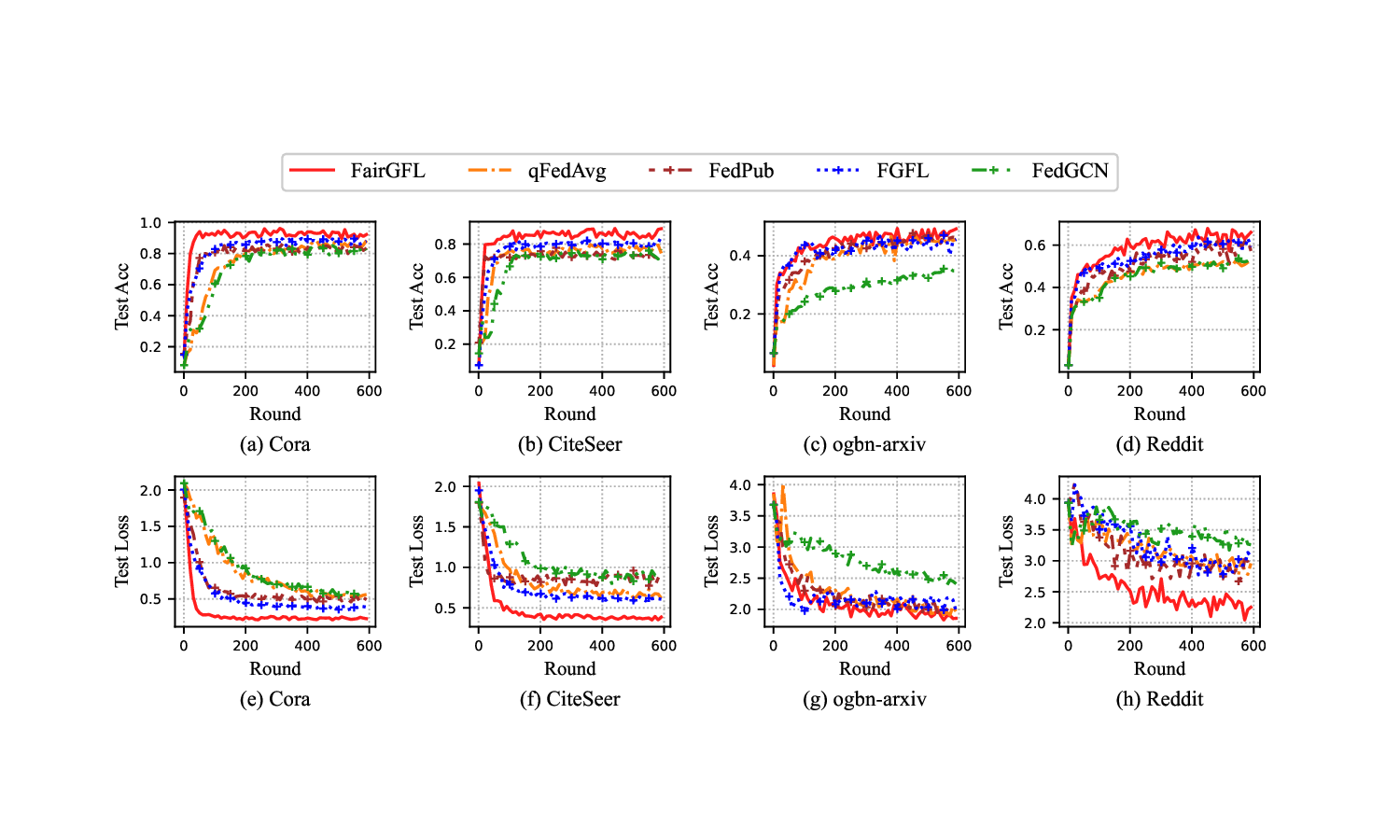}   
  \caption{The model utility of different algorithms on graph FL data (Cora, CiteSeer, Ogbn-arxiv, and Reddit). The model utility is evaluated by the test loss and accuracy.}\label{Exp:perf}
\end{figure*}

\begin{theorem}  \label{FairGFLConAnalysis}
  \textit{Assume Assumptions \ref{assume-Lsmooth}-\ref{assume-local} hold.
  Then, we have the following convergence result:}
  \begin{IEEEeqnarray}{cl}
      & \frac{1}{J} \sum_{j=1}^{J} \mathcal{A} \mathbb{E} \| \nabla F(w_{j}) \|^{2}  \leq  \frac{ F(w_{1}) - F(w^*) }{J} + \mathcal{B},   \nonumber
  \end{IEEEeqnarray}
  where $\mathcal{A} =  - \eta (E - 1/2) +  8 \eta E^{2} q_{\max} / K $, $\mathbb{E}[(q_{j,i} - {q_{i}})^2] \leq q_{max}$ and $ \beta = \frac{E-1}{1-\alpha}  - \frac{\alpha - \alpha^E}{(1-\alpha)^2} $
  and
  {\small
    \begin{IEEEeqnarray}{cl}
    &  \mathcal{B} = \frac{L(1+\lambda)}{2} \left( \lambda^2 E G_{l}^{2} + 4 G_{g}^2  \lambda^{2} E^2 +  \frac{2 \lambda q}{P} G_{l}^2 E^2  +  \frac{1}{K^2}  E G_{l}^{2}   \right) \nonumber \\
    & \quad +  \frac{G_{g}^2 L (1+\lambda)}{K}  E ( q_{\max} + 2E)  + 8 \eta E   q_{\max}  G_{g}^{2}   \nonumber  \\
    & \quad + \eta^2  \beta L^{3} (1+\lambda)  \left(   G_{l}^2  + 2 K   G_{g}^{2} \right)  \left( \frac{q_{\max} +2E }{K}  + 2 \lambda^{2} E  \right)      \nonumber  \\
    & \quad + \eta^{3} \beta  E    (4L^{2} q_{\max}  +  2L^{2} /K + \lambda^{2}  L^{2}) \left(  G_{l}^2  + 2 K   G_{g}^{2} \right)    .    \label{eq:final-B}
   \end{IEEEeqnarray} }
\end{theorem}
\begin{IEEEproof}
Please refer to Appendix \ref{sec:appCA}.
\end{IEEEproof}

Based on Theorem \ref{FairGFLConAnalysis}, we can qualitatively analyze the impact of heterogeneity and the convergence rate of FairGFL. To achieve the expected error smaller than $\epsilon$, we bound the number of rounds $J$ as follows
{\small
\begin{IEEEeqnarray}{cl}
    J & = \mathcal{O} \left(  \frac{E+E^2}{\epsilon} + \frac{E^2}{P\epsilon} + \frac{Eq_{max} }{K\epsilon}   +  \sqrt{ \frac{\beta E}{\epsilon} } \left(  1  + \sqrt{K} \right)  \right.   \nonumber \\
    &  \left. \quad + \left( \frac{\beta E}{\epsilon} \right)^{\frac{1}{3}} \left( \frac{1}{K^{1/3}}  +  E^{1/3} \right)  \left(  1  +  K^{1/3}  \right) \right).
\end{IEEEeqnarray}
}

\noindent By analyzing the rate of convergence, we can conclude that
\begin{itemize}[itemindent=5ex, leftmargin=0ex]
    \item The number of local iterations ($E$): The convergence result demonstrates that a larger value of $E$ allows clients to effectively exploit their local subgraphs, accelerating convergence. However, excessively mining local data leads to significant divergence and decreases the model utility. Hence, appropriate data mining with a smaller $E$ facilitates convergence, while excessive mining with a larger $E$ compromises model utility.
    \item The weights error ($q_{max}$): The term $q_{max}$ highlights that accurate estimation of overlapping ratios improves the model utility. With higher privacy budgets $\epsilon_{a}$ and $\epsilon_{b}$, the server can estimate overlapping ratios accurately, thus mitigating the impact of overlapping subgraphs on model performance.
\end{itemize}

\begin{figure*}[!t]
  \centering
  \includegraphics[scale=0.85, trim={55 67 56 68}, clip]{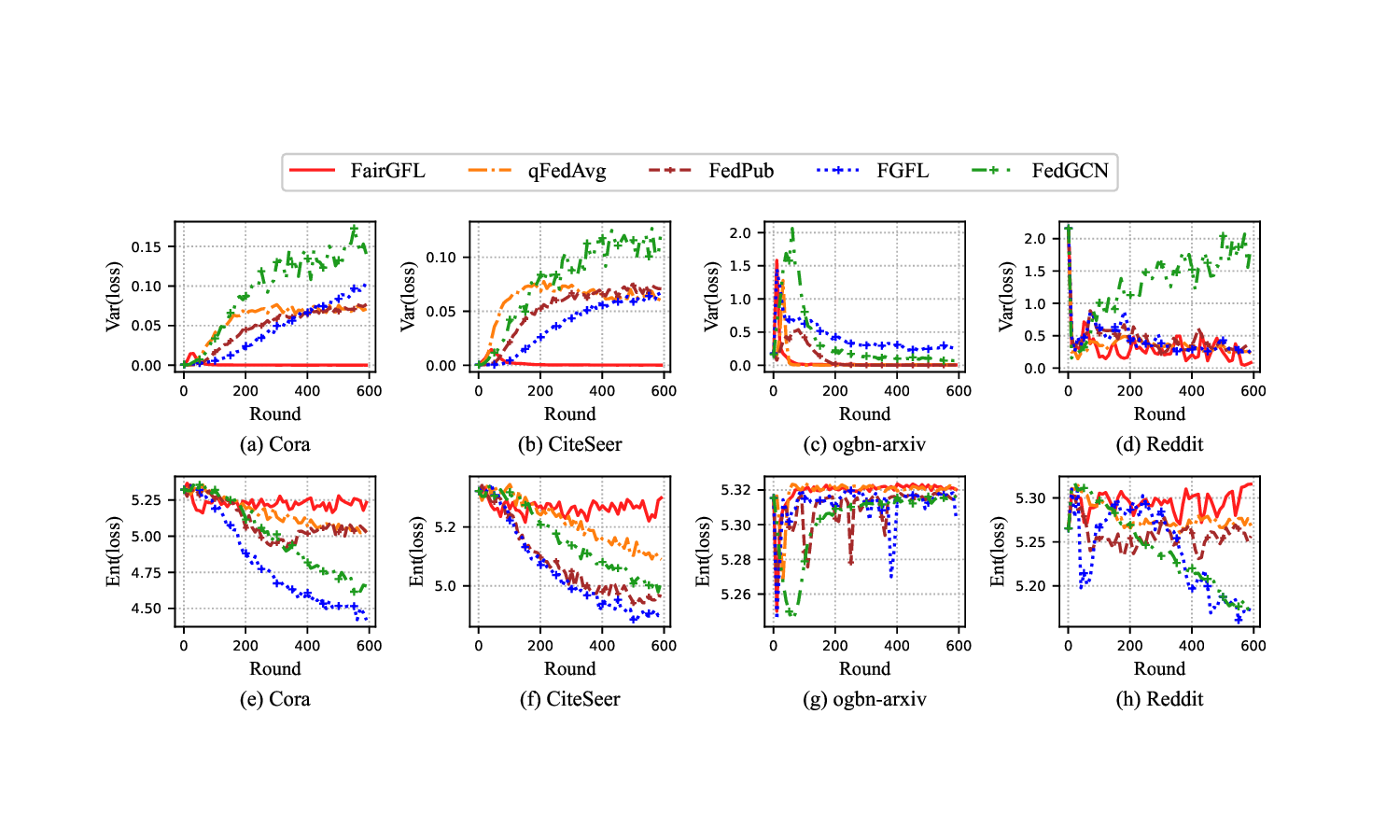}   
  \caption{The fairness (variance and entropy of local model utility) of different algorithms on graph FL data (Cora, CiteSeer, Ogbn-arxiv, and Reddit).}\label{Exp:fair}
\end{figure*}

\section{Experiments}\label{sec:experiments}

In this section, we conduct extensive experiments on four real-world graph data to demonstrate the effectiveness of FairGFL in terms of both fairness and model utility.


\subsection{Experimental Setup}\label{Subsec:Exp-Setup}

\textbf{Datasets.} We conducted experiments on the following four graph data. The statistical information is shown in Table~\ref{Tab:datasets}.
\begin{itemize}[itemindent=5ex, leftmargin=0ex]
  \item Cora~\cite{mccallum2000automating}: Cora consists of 2708 scientific publications and their corresponding citation network. Each publication is represented by a binary word vector indicating the absence (0) or presence (1) of words from a predefined dictionary.
  \item CiteSeer~\cite{giles1998citeseer}: CiteSeer consists of 3312 scientific publications and their corresponding citation network. Similar to Cora, each publication is described by a binary word vector based on word occurrences.
  \item Ogbn-arxiv~\cite{wang2020microsoft}: Ogbn-arxiv represents the citation network among Computer Science (CS) arXiv papers. It is a directed graph where each node corresponds to an arXiv paper, and directed links indicate citation relationships. Each paper is associated with a 128-dimensional feature vector obtained by averaging word embeddings from its title and abstract.
  \item Reddit~\cite{hamilton2017inductive}: Reddit originates from a large online discussion forum, where users post and comment on various topics. Each graph node represents a post and includes information such as the post's title, comments, and score. The node labels indicate the community to which the posts belong. Posts with common commenters are connected.
\end{itemize}

To simulate realistic federated scenarios with overlapping data, we divided the graph nodes into overlapping and non-overlapping subsets through a controlled proportion $r$. Non-overlapping nodes underwent non-IID distribution across clients via Dirichlet sampling parameterized by $\alpha_{dir}=0.5$.
The data volume of overlapping data for each label on each client is determined by sampling from a Dirichlet distribution with $\alpha_{dir}=0.8$, scaled to $\mathbf{N} / (r-\mathbf{N}r)$ times the total overlapping data, where $\mathbf{N}$ denotes the predefined overlap coefficient. Subsequently, for each label, the corresponding number of instances is randomly drawn without replacement from the overlapping dataset. In cases where the required sample size for a label surpasses its available instances, the sampling is capped at the total available data for that label.
Crucially, all client subgraphs preserved topological integrity by maintaining complete edge connectivity inherited from the global graph structure.

\begin{table}[!ht]
  \centering
  \caption{Basic Graph Information for four datasets.}\label{Tab:datasets}
  \begin{tabular}{cccccc}
    \toprule[0.5pt]
    Datasets  &   Nodes   &    Links   &    Features   &   Labels   \\   \midrule[0.1pt]
	Cora      &     2,708  &    10,858    &          1433        &      7      \\
	CiteSeer  &     3,327  &    9,464    &          3703        &      6      \\
    Reddit     &    232,965  &  114,615,892 &      602        &      41      \\
    Ogbn-arxiv  &    169,343  &  1,166,243   &         128        &      40      \\
    \bottomrule[0.5pt]
  \end{tabular}
\end{table}

\textbf{Default hyper-parameter settings.} We set the coupling weight $\alpha=0.8$, accumulated weight $\beta=0.5$, and privacy budgets $\epsilon_{a}=3$ for graph nodes and $\epsilon_{a}=1$ for links. Under this noise setting, FairGFL maintains accurate estimation of overlapping ratios.

\textbf{Model.}
In our experiments, we took the representative two-layer Graph Convolutional Networks (GCN) as our model \cite{kipf2016semi} and cross-entropy loss as the loss function.

\textbf{Comparison algorithms.}
We compared FairGFL  with the baseline FedGCN~\cite{yao2024fedgcn}, and three representative algorithms:
\begin{itemize}[itemindent=5ex, leftmargin=0ex]
    \item \emph{Q-FedAvg.} Q-FedAvg~\cite{lifair} modifies the objective function by magnifying the maximum loss among all clients. The objective function can be expressed as follows
    \begin{IEEEeqnarray}{rl}
      \min_{w} F(w) = \sum_{i=1}^{m} \frac{p_i}{q+1} F_{i}^{q+1}(w),     \nonumber
    \end{IEEEeqnarray}
    where $q>0$ is a constant and $p_i$ defines the weight.

    \item \emph{FGFL.} FGFL~\cite{2023Towards} enhances accuracy-fairness trade-off by designing a dual-incentive mechanism that integrates both model gradients and payoff allocation. FGFL quantifies agent contributions through a valuation function based on gradient alignment and graph diversity. Then, it allocates sparsified gradients as immediate rewards and distributes payoff with penalties for harmful agents and compensation for those with delayed contributions.

    \item \emph{FedPub.} FedPub~\cite{baek2023personalized} aims to address the impact of heterogeneity among subgraph data by learning multiple personalized models. Specifically, FedPub introduces an innovative for computing similarities based on the disparity among outputs given the same input data. To achieve efficient communication, each client uploads partial information based on personalized sparse masks.
\end{itemize}

\textbf{Metrics.}

We evaluate FairGFL in terms of both model utility and fairness. Model utility is measured using the accuracy and loss of the global model $w$ on a held-out global test set.
For fairness evaluation, we employ entropy and variance computed from the losses of the global model $w$ over all local training datasets $\mathcal{D}_{i} \mid i \in [P]$. Different fairness notions require specific quantitative metrics tailored to their respective objectives. These metrics are chosen because they directly measure performance consistency across clients, which aligns with our objective.
Generally, a higher entropy or a lower variance indicates a more equitable performance distribution across clients, reflecting better cross-client fairness. The metrics are formally defined as follows:

\begin{IEEEeqnarray}{rl}
  \text{Ent}\text{(loss)} =  & - \sum_{i=1}^{P}  \frac{F(w, \mathcal{D}_{i})}{||\overrightarrow{\mathbf{F}}(w)||_{1}} \log \frac{F(w, \mathcal{D}_{i})}{||\overrightarrow{\mathbf{F}}(w)||_{1}},  \nonumber    \\
  \text{Var}\text{(loss)} =  & \frac{1}{P} \sum_{i=1}^{P} \left(F(w, \mathcal{D}_{i}) - \frac{1}{P} \sum_{p=1}^{P} F(w, \mathcal{D}_{p}) \right)^2,  \nonumber
\end{IEEEeqnarray}
where $||\overrightarrow{\mathbf{F}}(w)||_{1} = \sum_{i=1}^{P}F(w, \mathcal{D}_{i}) $.

\begin{table*}[!ht]
  \centering
  \caption{Normalized overlapping ratios of graph nodes on Cora and Citeseer datasets.}\label{Tab:overlap_rate}
  \begin{tabular}{ccccccccccccccc}
	\toprule[0.5pt]
	Dataset & Privacy budget &
    \multicolumn{4}{c}{True overlapping graph node ratios} &  \multirow{2}*{} & \multicolumn{4}{c}{Estimated overlapping graph node ratios}  \\ \midrule[0.2pt]
    \multirow{3}*{Cora} & \multirow{3}*{$\epsilon_a=3$} &  0.0521 & 0.0352& 0.0289& 0.0676  & & 0.0367& 0.0421& 0.0312& 0.0608 \\
    &  &  0.0315& 0.0423& 0.0187& 0.0489 &  & 0.0231& 0.0356& 0.0238& 0.0375  \\
    &  &  0 & 0& 0 & 0 & & 0.0058& 0.0083& 0.0091& 0.0072 \\ \midrule[0.1pt]
    \multirow{3}*{CiteSeer} & \multirow{3}*{$\epsilon_a=3$} & 0.0475 & 0.0312 & 0.0458 & 0.0364 & & 0.0452 & 0.0256 & 0.0398 & 0.0387  \\
    &  &  0.0289& 0.0197& 0.0341& 0.0512 & & 0.0323 & 0.0169 & 0.0356 & 0.0531  \\
    &  & 0 & 0& 0 & 0 & &  2.8512e-10 & 0.0042 & 3.2732e-12 & 7.8254e-08 \\
	\bottomrule[0.5pt]
  \end{tabular}
\end{table*}

\begin{figure}[!t]
  \centering
  \includegraphics[scale=0.7, trim={13 42 0 53}, clip]{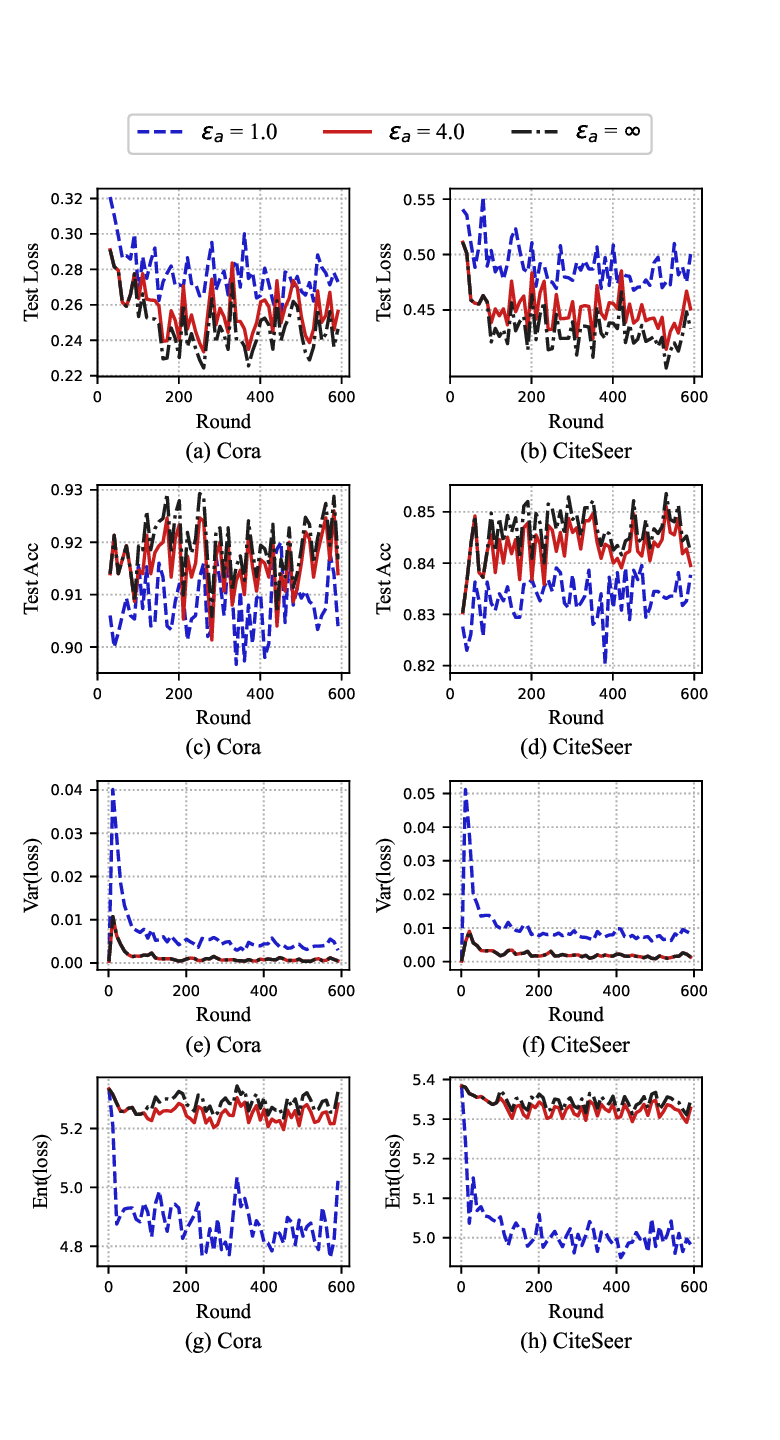}   
  \caption{The model utility and disparities (variance and entropy of local model utility) of FairGFL under different privacy budgets ($\epsilon_{a}$=1.0, 4.0 and $\infty$) on the Cora and CiteSeer. Privacy budget $\epsilon_{a}=\infty$ means that LDP mechanism is not employed.}\label{Exp:epsilon}
\end{figure}

\subsection{Experimental Results and Analysis}\label{Subsec:Exp-results}

To demonstrate the superiority of FairGFL in the sense of improving the tradeoff between model utility and fairness,  and explore the impact of certain parameters, the following three groups of experiments were conducted.
\begin{enumerate}
  \item The first group compared FairGFL to baseline algorithms in terms of both model utility and fairness on all four datasets and analyzed the additional communication costs.
  \item The second group explored the impact of different privacy budgets on the model utility and fairness of FairGFL.
  FairGFL with respect to the varying overlapping ratios in terms of both model utility and fairness.
\end{enumerate}

\subsubsection{Comparison with different algorithms}

\textit{Small synthetic FL graph data (Cora and CiteSeer).}
\figurename~\ref{Exp:perf}(a) and \figurename~\ref{Exp:perf}(b), as well as \figurename~\ref{Exp:perf}(e) and \figurename~\ref{Exp:perf}(f) depict the model performance of different algorithms on Cora and CiteSeer. As shown, \textbf{FairGFL consistently exhibits the fastest convergence to the highest prediction accuracy}. Such advantage mainly arises from the proposed composite losses optimization in Section~\ref{Sec:method-sec4}.
Furthermore, \figurename \ref{Exp:fair}(a) and \figurename \ref{Exp:fair}(b), as well as \figurename \ref{Exp:fair}(e) and \figurename \ref{Exp:fair}(f) demonstrate that \textbf{FairGFL achieves the smallest variance and a larger entropy, indicating the desired cross-client fairness}. This can be attributed to the proposed fairness-aware aggregation strategy in Section~\ref{Sec:method-sec3}. In contrast, FedPub exhibits the poorest performance, with significant variation in model utility across clients. This is due to the fact that FedPub aims to learn multiple personalized models and data heterogeneity disparates the performance across clients. Additionally, FedGCN also exhibits poorer performance, which highlights the effectiveness of these fairness-aware algorithms.

\begin{figure}[!t]
  \centering
  \includegraphics[scale=0.76, trim={13 40 0 13}, clip]{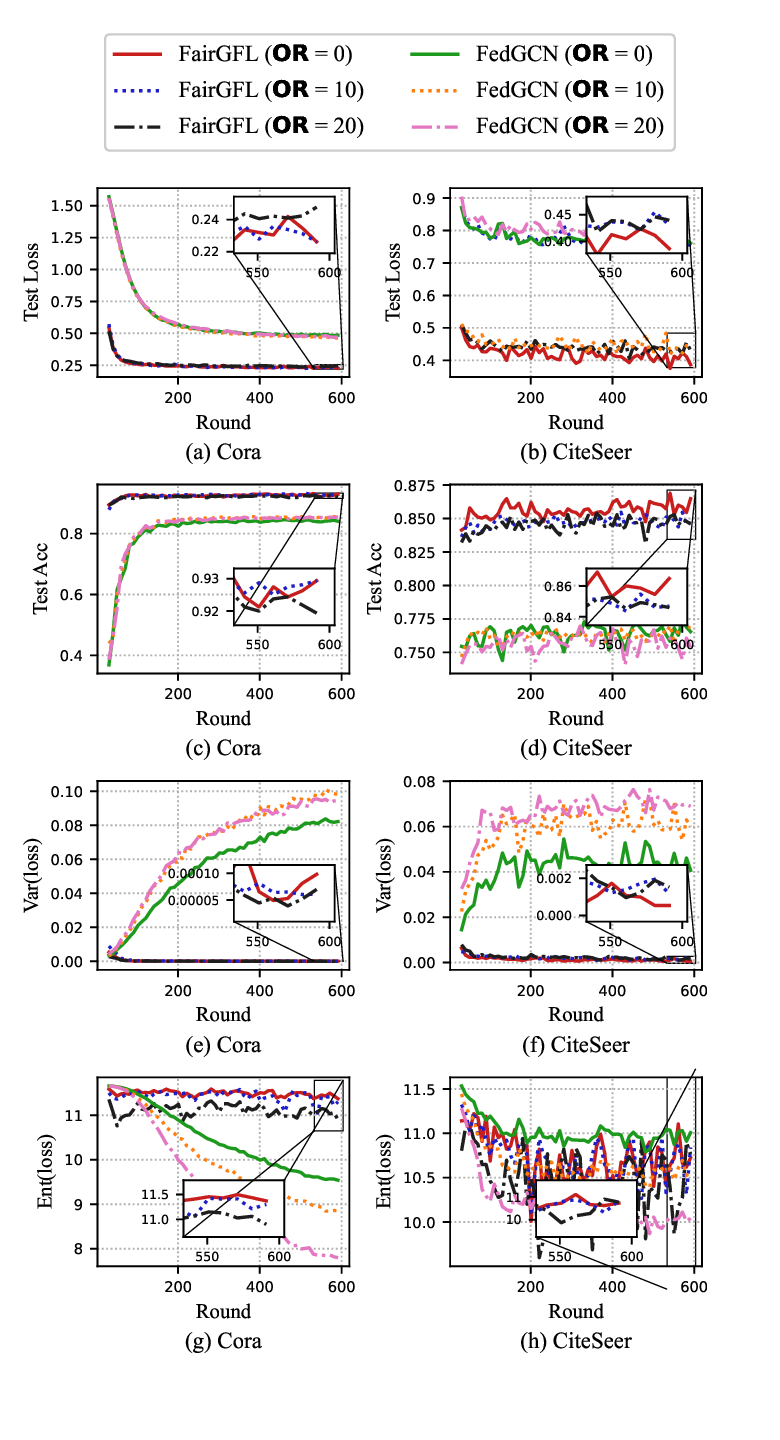}   
  \caption{The model utility and disparities (variance and entropy of local model utility) of FairGFL and the baseline under different overlapping ratios (\%) on Cora and CiteSeer.}\label{Exp:overrate}
\end{figure}

\textit{Large synthetic FL graph data (Reddit and Ogbn-arxiv).}
\figurename~\ref{Exp:perf}(c) and \figurename~\ref{Exp:perf}(d), as well as \figurename~\ref{Exp:perf}(g) and \figurename~\ref{Exp:perf}(h) shows the model performance of different algorithms on Reddit and Ogbn-arxiv. Similar to the results obtained on small graph data, FairGFL consistently achieves the fastest convergence and highest prediction accuracy, indicating its excellent performance. As depicted in \figurename~\ref{Exp:fair}(c) and \figurename~\ref{Exp:fair}(d), as well as \figurename~\ref{Exp:fair}(g) and \figurename~\ref{Exp:fair}(h), FairGFL exhibits a lower variance and a larger entropy, indicating a fairer performance. Notably, FGFL performs worst since FGFL prioritizes high contributors with full model updates. This unequal resource allocation directly amplifies the divergent performance of the global model across clients. It enforces contribution-based fairness at the cost of cross-client fairness.

\begin{table}[!ht]
  \centering
  \caption{Communication bytes for model updates and privacy-preserving subgraphs, and the ratio of additional communication costs (i.e. the ratio of communication bytes of both node features and links to that of model).}\label{Tab:commu}
  \begin{tabular}{cccccc}
    \toprule[0.5pt]
    Datasets  &   Model   &    Features  &   Links & Ratio  \\   \midrule[0.1pt]
	Cora      &  1475612  &    25600     &     512  &  1.77\%            \\
	CiteSeer  &  1899544  &   76800      &  512     &  4.07\%             \\
    Reddit    &  329892   &  20480       &  4096    &  7.45\%       \\
    Ogbn-arxiv &  346272  &   16384      &  4096    &  5.91\%          \\
    \bottomrule[0.5pt]
  \end{tabular}
\end{table}

\textit{Communication analysis.}
FairGFL necessitates clients to upload privacy-preserving subgraphs for the estimation of overlapping ratios, thereby increasing the communication cost. To quantize the extra communication burden on each client, we count the communication cost of transmitting model updates and the subgraphs, as depicted in Table \ref{Tab:commu}. Experimental results indicate a maximum of 8\% additional communication cost. Consequently, \textbf{FairGFL significantly enhances both fairness and model utility while maintaining a manageable level of extra communication overhead.}

\subsubsection{The impacts of different privacy budgets}

Table~\ref{Tab:overlap_rate} shows the normalized overlapping graph node ratios on Cora and CiteSeer data under the privacy budget $\epsilon_a=3$. The number of clients is set to 40 and we display eight overlapping clients (rows 2-3 and rows 5-6) and four non-overlapping clients (rows 4 and 7) due to the limitation of space. \figurename~\ref{Exp:epsilon} depicts the model performance of FairGFL under different privacy budgets on Cora and CiteSeer data.

We draw two conclusions from Table~\ref{Tab:overlap_rate} and \figurename~\ref{Exp:epsilon}.
\textbf{On the one hand, FairGFL can accurately estimate the overlapping ratios.}
By comparing the normalized true overlapping graph node ratios and normalized estimated overlapping graph node ratios, we can see that it is easy to distinguish between overlapping and non-overlapping clients and identify the level of overlapping ratios.
\textbf{On the other hand, the model utility improves with the increasing privacy budget}, indicating the tradeoff between the privacy and model utility, as shown in \figurename~\ref{Exp:epsilon}. Moreover, As the privacy budget increases, the model fairness presents an obvious improvement in terms of variance and entropy. This is reasonable since a larger privacy budget can lead to a more accurate estimation of overlapping ratios, thereby a more effective fairness-aware aggregation.

\subsubsection{The impacts of different overlapping ratios}

\figurename~\ref{Exp:overrate} presents the model performance under different overlapping ratios on Cora and CiteSeer data. As depicted in \figurename~\ref{Exp:overrate}(a)-\figurename~\ref{Exp:overrate}(d), a larger overlapping ratio leads to a higher test loss and lower test accuracy, which can be explained by the overfitting of model on the overlapping data (same as the analysis in Section \ref{sec:motivation}). We can conclude that
\textbf{(i) FairGFL exhibits a superior model performance} in contrast to FedGCN, indicating its effectiveness in eliminating overfitting and improving model utility.
\textbf{(ii) as the overlapping ratio increases, the performance inconsistency among clients becomes more obvious}, as demonstrated in \figurename~\ref{Exp:overrate}(e)-\figurename~\ref{Exp:overrate}(h), consistent with the conclusion in Section~\ref{sec:motivation}.  \textbf{(iii) FairGFL exhibits a more robust performance with respect to the varying overlapping ratios than FedGCN}. The result illustrates that the fairness-aware aggregation strategy in Section~\ref{Sec:method-sec3} effectively mitigates the impact of overlapping ratios benefiting from the accurate estimation of overlapping ratios (Section~\ref{Sec:method-sec2}) and well-crafted weights (Section~\ref{Sec:method-sec3}).

\section{Conclusion}\label{sec:conclusion}

In this paper, we propose FairGFL, a novel fairness-aware subgraph federated learning algorithm, targeting to enhance fairness and model utility. FairGFL achieves these improvements through three key contributions.
First, FairGFL ensures the privacy of clients' raw subgraphs by utilizing encoders and noval LDP techniques.
Second, FairGFL enhances fairness by accurately estimating the overlapping ratios of different clients based on the sanitized mini-batch subgraphs.
Third, FairGFL improves the tradeoff between fairness and model utility by incorporating a well-crafted regularizer into the federated composite losses.
We conducted experiments on four real-world graph datasets, and the results demonstrate the effectiveness of FairGFL in enhancing fairness and maintaining model utility.

\bibliographystyle{IEEEtranN}
{\footnotesize \bibliography{ref}}

\vspace{-1.9em}
\begin{IEEEbiography}[{\includegraphics[width=1in,height=1.25in, clip,keepaspectratio]{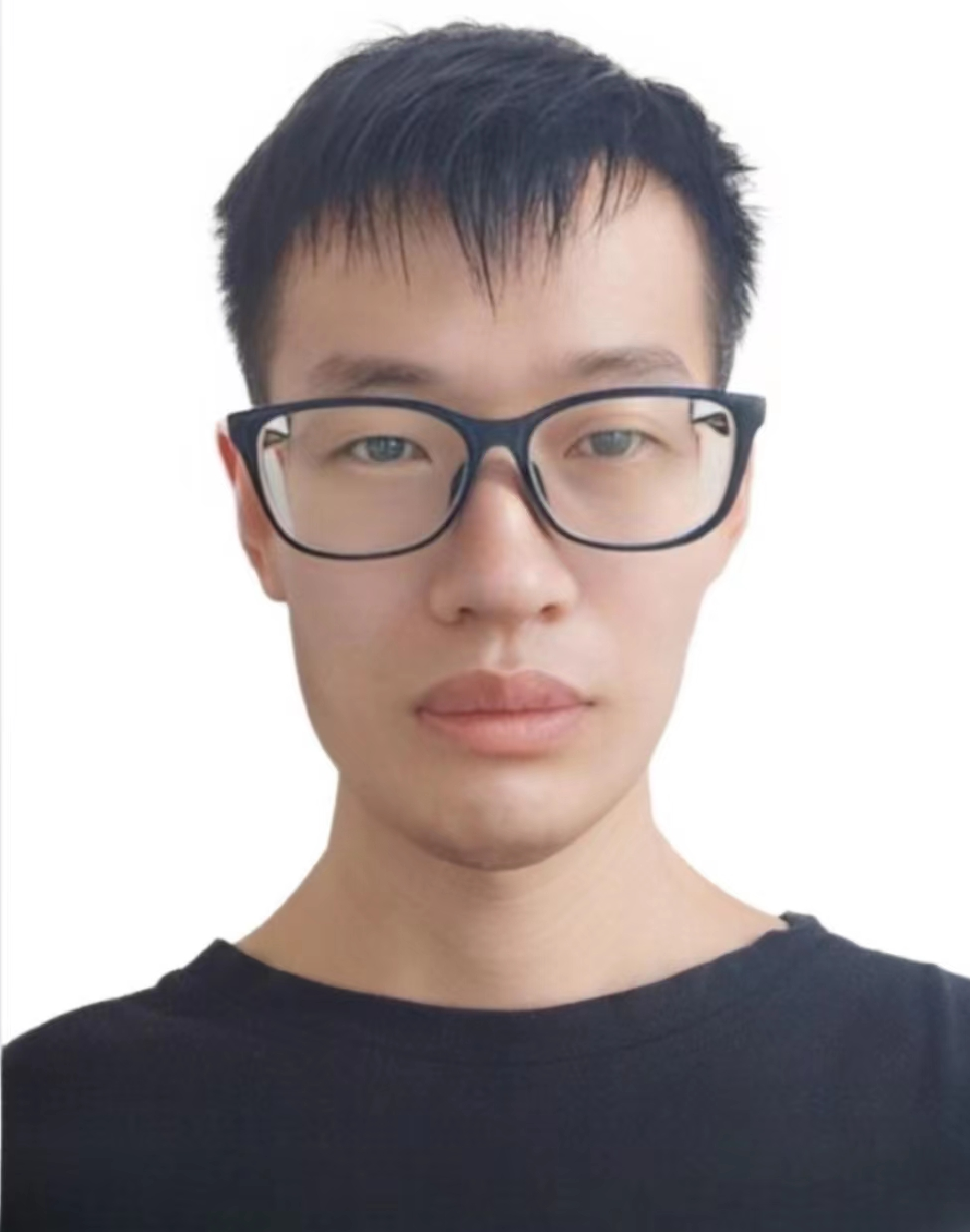}}]{Zihao Zhou} received the BSc degree in the School of Mathematics and Statistics from Xi'an Jiaotong University (XJTU), China, in 2020. He is currently working toward the PhD degree in the School of Mathematics and Statistics and is a member of the National Engineering Laboratory for Big Data Analytics (NEL-BDA), both from XJTU, China. His research interests include federated learning and edge-cloud intelligence.
\end{IEEEbiography}
\vspace{-1.9em}

\vspace{-1.9em}
\begin{IEEEbiography}[{\includegraphics[width=1in,height=1.25in, clip,keepaspectratio]{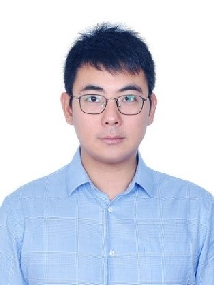}}] {Shusen Yang} (Senior Member, IEEE) received the PhD degree in computing from Imperial College London, in 2014. He is currently a professor, and the deputy director of the National Engineering Laboratory for Big Data Analytics and the Ministry of Education (MoE) Key Laboratory for Intelligent Networks and Network Security, Xi’an Jiaotong University (XJTU), Xi’an, China. Before joining XJTU, he worked as a lecturer (assistant professor) with the University of Liverpool, Liverpool, U.K., from 2015 to 2016, and a research associate with the Intel Collaborative Research Institute (ICRI) on sustainable connected cities, from 2013 to 2014. His research interests include distributed systems and data sciences, and their applications in industrial scenarios, including data-driven network algorithms, distributed machine learning, edge-cloud intelligence, industrial internet, and industrial intelligence. He is a member of the ACM. He is also a DAMO Academy Young fellow and an honorary research fellow with Imperial College London.
\end{IEEEbiography}
\vspace{-1.9em}

\vspace{-1.9em}
\begin{IEEEbiography}[{\includegraphics[width=1in,height=1.25in, clip,keepaspectratio]{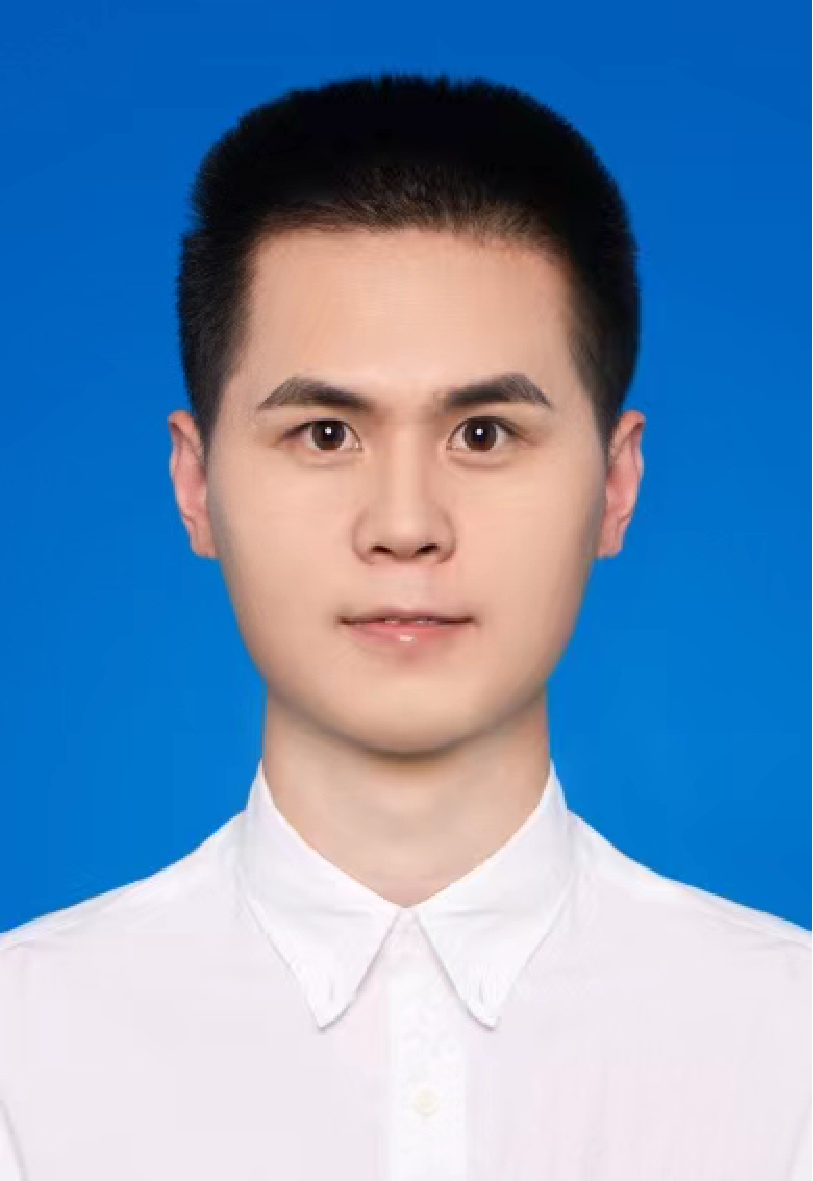}}] {Fangyuan Zhao}  received the BSc degree from the School of Mathematics and Statistics, Xi’an Jiaotong University (XJTU), China, in 2018. He is currently working toward the PhD degree in the School of Computer Science and Technology and is a member of the National Engineering Laboratory for Big Data Analytics (NEL-BDA), both from XJTU, China. His research interests include differential privacy, privacy-preserving machine learning, and federated learning.
\end{IEEEbiography}
\vspace{-1.9em}

\vspace{-1.9em}
\begin{IEEEbiography}[{\includegraphics[width=1in,height=1.25in,clip,keepaspectratio]{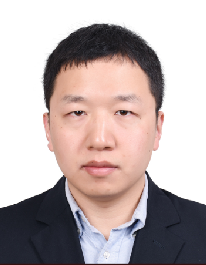}}]{Xuebin Ren} (Member, IEEE) received the PhD degree from the Department of Computer Science and Technology, Xi’an Jiaotong University, China, in 2017. He is currently an associate professor with the School of Computer Science and Technology and a member of the National Engineering Laboratory for Big Data Analytics (NEL-BDA), both with Xi’an Jiaotong University. He has been a visiting PhD student with the Department of Computing, Imperial College London, from 2016 to 2017. His research interests include data privacy protection, federated learning, and privacy-preserving machine learning. He is a member of the ACM.
\end{IEEEbiography}
\vspace{-1.9em}

\appendices

\clearpage

\section{Analysis of Overlaps on Fairness}\label{sec:moti}

To provide an in-depth insight into the impact of overlapping data on fairness for non-graph data, we theoretically analyze the average Euclidean distance between the loss value of each local model and the average loss value.
\begin{IEEEeqnarray}{cl}\label{moti-eq}
    & \quad \frac{1}{P} \sum_{i=1}^{P} || F(w, \mathcal{D}_{i}) - \frac{1}{P} \sum_{k=1}^{P} F(w, \mathcal{D}_{k}) ||   \nonumber  \\
    & \leq \frac{1}{P^2} \sum_{i=1}^{P} \sum_{k=1}^{P} || F(w, \mathcal{D}_{i}) -  F(w, \mathcal{D}_{k}) ||      \label{moti-eq-1}  \\
    & \leq \frac{1}{n^2} \left( \sum_{d_{k_o^1}^{o} \in \mathcal{D}_{o}'} \sum_{d_{k_o^2}^{o} \in \mathcal{D}_{o}'} ||  F(w, d_{k_o^1}^{o}) - F(w, d_{k_o^2}^{o}) || + \right.   \nonumber  \\
    & \quad  \left.  \sum_{d_{k_o}^{o} \in \mathcal{D}_{o}'} \sum_{d_{k_n}^{n} \in \mathcal{D}_{n}'} ||  F(w, d_{k_o}^{o}) - F(w, d_{k_n}^{n}) ||   + \right.  \nonumber  \\
    & \quad  \left.  \sum_{d_{k_n^1}^{n} \in \mathcal{D}_{n}'} \sum_{d_{k_n^2}^{n} \in \mathcal{D}_{n}'} ||  F(w, d_{k_n^1}^{n}) - F(w, d_{k_n^2}^{n}) || \right)  \nonumber \\
    & \leq \frac{1}{n^2} \left( \sum_{d_{k_o^1}^{o} \in \mathcal{D}_{o}'} \sum_{d_{k_o^2}^{o} \in \mathcal{D}_{o}'}  L_{d} \|  d_{k_o^1}^{o} - d_{k_o^2}^{o} \| + \right.   \nonumber  \\
    & \quad  \left.  \sum_{d_{k_o}^{o} \in \mathcal{D}_{o}'} \sum_{d_{k_n}^{n} \in \mathcal{D}_{n}'}  L_{d} \|  d_{k_o}^{o} - d_{k_n}^{n} \|   + \right.  \nonumber  \\
    & \quad  \left.  \sum_{d_{k_n^1}^{n} \in \mathcal{D}_{n}'} \sum_{d_{k_n^2}^{n} \in \mathcal{D}_{n}'}  L_{d} \|  d_{k_n^1}^{n} - d_{k_n^2}^{n} \| \right).   \label{moti-eq-2}
\end{IEEEeqnarray}
Eq.\eqref{moti-eq-1} is derived by triangle inequality and Eq.~\eqref{moti-eq-2}  is derived by Eq.~\eqref{moti-eq:similardata}.

\section{Local Differential Privacy}\label{sec:appDP}

\begin{IEEEproof}[Proof of Theorem \ref{TH:nodenoise}]\label{ap:proof:nodenoise}
  Given two arbitrary distinct graph node feature values $x_{i,k}, \tilde{x}_{i,k} \in [x_{\min}, x_{\max}]$ ($x_{i,k} \neq \tilde{x}_{i,k}$), where $x_{\min}$ and $x_{\max}$ represent the minimum and maximum values of the graph node feature $x_{i,k}$ respectively. Then, given any output $s_{i,k} = \frac{p_{i,k}}{p}, p_{i,k} \in [p]$ from $\mathcal{M}_{N}$, we can state that
  \begin{IEEEeqnarray}{rl}
    & \frac{ \Pr [\mathcal{M}_{N} (x_{i,k}) = s_{i,k} ] }{ \Pr [\mathcal{M}_{N} (\tilde{x}_{i,k}) = s_{i,k} ] }  =
    \frac{e^{\epsilon_a ( 1 - \frac{1}{p}\lfloor  p |\frac{x_{i,k}-x_{\min}}{x_{\max} - x_{\min}} - \frac{p_{i,k}}{p} | \rfloor ) }}{e^{\epsilon_a ( 1 - \frac{1}{p}\lfloor  p |\frac{\tilde{x}_{i,k}-x_{\min}}{x_{\max} - x_{\min}} - \frac{p_{i,k}}{p} | \rfloor ) }}  \nonumber  \\
     & =
    e^{\epsilon_a ( \frac{1}{p}\lfloor  p |\frac{\tilde{x}_{i,k}-x_{\min}}{x_{\max} - x_{\min}} - \frac{p_{i,k}}{p} | \rfloor - \frac{1}{p}\lfloor  p |\frac{x_{i,k}-x_{\min}}{x_{\max} - x_{\min}} - \frac{p_{i,k}}{p} | \rfloor ) } \leq e^{\epsilon_a}.  \nonumber
  \end{IEEEeqnarray}
\end{IEEEproof}

\section{Overlapping Ratio Estimation}\label{sec:appORE}
Before presenting the proof of Theorem~\ref{TH:node-over-rates}, we introduce the following technical lemma along with its proof.
\begin{lemma}\label{Lemma:node-over-rate}
Consider two clients, denoted as client $i$ and client $k$, each with $n_i$ and $n_k$ graph nodes respectively. If client $k$ randomly samples $b_k$ graph nodes and client $i$ has a proportion $p_{i,k} \in [0, 1]$ of graph nodes overlapping with the $b_k$ samples, then client $i$ has an expected proportion $p$ of graph nodes overlapping with client $k$, where
\begin{IEEEeqnarray}{c}
  p = \min \{ 1, \frac{p_{i,k} n_k}{b_k} \}.
\end{IEEEeqnarray}
\end{lemma}
\begin{IEEEproof}
    The proof follows from the combination of the definition of expectation in probability theory and the properties of random sampling.
\end{IEEEproof}

\begin{IEEEproof}[Proof of Theorem~\ref{TH:node-over-rates}]\label{ap:TH:node-over-rates}
    Based on Lemma~\ref{Lemma:node-over-rate}, since client $i$ has a proportion $p_{i,k} \in [0, 1]$ of samples overlapping with the $b_k$ samples of client $k$, then client $i$ may have a proportion $\frac{p_{i,k} n_k}{b_k}$ of samples overlapping with client $k$. This is equivalent to stating that client $k$ may have a proportion $\frac{p_{i,k} n_k b_i}{n_k b_k}$ of graph nodes overlapping with client $i$. By applying Lemma~\ref{Lemma:node-over-rate} again, we can deduce that client $k$ may have a proportion $\frac{p_{i,k} n_i^2}{n_k b_k}$ of graph nodes overlapping with client $i$, which aligns with the conclusion.
\end{IEEEproof}

\begin{IEEEproof}[Proof of Theorem \ref{TH:link-over-rates}]\label{ap:TH:link-over-rates}
  Since the overlapping graph nodes are randomly sampled, the overlapping links are uniformly distributed across these graph nodes. Consequently, the link overlapping ratio on mini-batch overlapping graph nodes is equal to the ratio among all the overlapping graph nodes. Specifically, the link overlapping ratio on mini-batch overlapping graph nodes can be expressed as $\frac{ \mathbf{T}_{i,k}^{sub} }{ p_{i,k}^2 }$. Similar to Lemma \ref{Lemma:node-over-rate}, by combining the definition of expectation in probability theory and random sampling, we can deduce that the overall link overlapping ratio is $\frac{ \mathbf{T}_{i,k}^{sub} N_{k}^{2} }{ b_i^2 }$.
\end{IEEEproof}

We consider various scenarios, including unreasonable ones, and analyze a set of weights to address the impact of imbalanced overlaps. Drawing inspiration from these weights, we propose an aggregation strategy that effectively enhances fairness, performing well in both common and uncommon scenarios. The analysis of the weight set is presented as follows.

\section{Fairness-aware Aggregation}\label{sec:appFA}

In this section, we consider unreasonable scenarios and analyze the setting of weights to mitigate the impact of imbalanced overlaps. Drawing inspiration from these weights, we propose an aggregation strategy to enhance fairness which also performs effectively in realistic scenarios. The analysis of weight configuration is presented as follows.

As shown in Eq.~\eqref{eq:fairlossRe}, when setting uniform weights across clients, it may cause nodes and links with higher overlapping ratios to dominate the model updates. This results in higher model utility for local models trained on data with significant overlaps, thereby introducing fairness issues. To achieve cross-client fairness, FairGFL adaptively adjusts the weights, effectively discounting the influence of highly overlapping data. This ensures a more balanced contribution across diverse data and enhances cross-client fairness. Then, we consider a special case and analyze the optimal weight.

\begin{theorem}\label{TH:final-weight}
    Assume that a pair of clients share the same subgraphs if they share a graph node. In this setting, to eliminate the impact of overlapping subgraphs and achieve fairness, the optimal weight for the $i$-th client with an overall overlapping ratio is
    \begin{IEEEeqnarray}{c}
        \frac{1}{ 1+\mathbf{O}_{i} }.
    \end{IEEEeqnarray}
\end{theorem}

\begin{IEEEproof}\label{Appendix:proof-final-weight}   
Consider a scenario where there are $P$ clients, each having their local subgraphs denoted as $\mathcal{G}_{i} = \{\mathcal{V}_{i}, \mathcal{E}_{i}, \mathbf{X}_{i}\}$. These subgraphs can be categorized into two distinct types: overlapping subgraphs and non-overlapping subgraphs.

The set of overlapping subgraphs can be defined as $\mathcal{G}^{o} = \{\mathcal{G}_{1}^{o}, \ldots, \mathcal{G}_{n_1'}^{o}\}$, where $n_1'$ represents the number of overlapping subgraphs without repetition. Given that the subgraph $\mathcal{G}_{i}^{o}$ on the $i$-th client overlaps $1+\mathbf{O}_{i}$ ($\mathbf{O}_{i} > 0$) times, the weight $q_{i}^{o}$ of the subgraph $\mathcal{G}_{i}^{o}$ can be expressed as
{\footnotesize
\begin{IEEEeqnarray}{cl}
  q_{i}^{o} & = \frac{1}{1+\mathbf{O}_{i}}.
\end{IEEEeqnarray}
}
Similarly, the set of non-overlapping subgraphs can be defined as $\mathcal{G}^{n} = \{\mathcal{G}_{1}^{n}, \ldots, \mathcal{G}_{n_2'}\}$, where $n_2'$ represents the number of non-overlapping subgraphs with $\mathbf{O}_{i} = 0$. Thus, the weight $q_{i}^{n}$ of the subgraph $\mathcal{G}_{i}^{n}$ can be expressed as
{\footnotesize
\begin{IEEEeqnarray}{cl}
  q_{i}^{n} = \frac{1}{1+\mathbf{O}_{i}} & = 1.
\end{IEEEeqnarray}
}

We define a set $\mathcal{G}$ that includes both overlapping and non-overlapping data without duplication, i.e., $\mathcal{G} = \{\mathcal{G}_{1}^{o}, \ldots, \mathcal{G}_{n_1'}^{o}, \mathcal{G}_{1}^{n}, \ldots, \mathcal{G}_{n_2'}\}$.

Then, by setting the weight of the $i$-th client as $\frac{1}{ 1 + \mathbf{O}_{i} }$ $(i=1, \ldots, P)$, the composite empirical loss function can be expressed as:
{\footnotesize
\begin{IEEEeqnarray}{cl}
    & \quad \sum_{i=1}^{P} \frac{1}{1+\mathbf{O}_{i}} F(w, \mathcal{G}_{i})   \nonumber  \\
    & = \sum_{i=1}^{n_1'} q_{i}^{o} {(1+\mathbf{O}_{i})} F(w, \mathcal{G}_{i}^{o}) + \sum_{i=1}^{n_2'} q_{i}^{n} {(1+\mathbf{O}_{i})} F(w, \mathcal{G}_{i}^{n})    \nonumber    \\
    & = \sum_{i=1}^{n_1'} F(w, \mathcal{G}_{i}^{o}) + \sum_{i=1}^{n_2'} F(w, \mathcal{G}_{i}^{n})  \nonumber  \\
    & = \sum_{\mathcal{G}_{i} \in \mathcal{G}} F(w, \mathcal{G}_{i}) .    \nonumber
\end{IEEEeqnarray}}
From this expression, we can observe that the weights of both overlapping and non-overlapping data are consistent. Therefore, the algorithm ensures fairness in terms of overlap.

\end{IEEEproof}

\section{Convergence Analysis}\label{sec:appCA}

\begin{IEEEproof}[Proof of Theorem \ref{FairGFLConAnalysis}]
We first analyze the convergence rate of one round of FairGFL. Then based on the results, we analyze the convergence of FairGFL.


The set of sampled clients is denoted as $\mathcal{S}_{j}^{c} \subseteq [P], |\mathcal{S}_{j}^{c}| = K$.
Based on the property of client-level unbiased gradients in Assumption~\ref{assume-local}, we have
{\footnotesize
\begin{IEEEeqnarray}{Cl}\label{aproof:beforeExp-prior}
  \mathbb{E} & \left[  \frac{1}{K} \sum_{i=1}^{K} q_{j,i} \sum_{e=0}^{E-1} \nabla f(w_{j,i}^{e}, \xi_{j,i}^{e})  \right]  = \mathbb{E}_{ \mathcal{S}^{c} } \left[ \frac{1}{K} \sum_{i=1}^{K} q_{i}' \sum_{e=0}^{E-1} \nabla F_{i} (w_{j,i}^{e})  \right]  \nonumber   \\
  & =  \sum_{i=1}^{P} \frac{ \binom{P-1}{K-1} }{ \binom{P}{K} } \frac{1}{K} q_{i}' \sum_{e=0}^{E-1} \nabla F_{i} (w_{j,i}^{e})  =  \frac{1}{P}  \sum_{i=1}^{P}  q_{i}' \sum_{e=0}^{E-1} \nabla F_{i} (w_{j,i}^{e}).
\end{IEEEeqnarray}
}

In the $j$-th round, the aggregated rule is as follows
{\footnotesize
\begin{IEEEeqnarray}{cl}\label{aproof:agg-rule}
    w_{j+1} & = w_{j} - \left( \frac{1}{K} \sum_{i=1}^{K} \frac{1}{ 1+\mathbf{O}_{j,i} } \Delta w_{j,i} + \lambda \Delta w_{j, i_{m}^{K}}  \right)   \nonumber   \\
    & = w_{j} - \eta \left(\frac{1}{K} \sum_{i=1}^{K} q_{j,i} \sum_{e=0}^{E-1} \nabla f(w_{j,i}^{e}, \xi_{j,i}^{e})    \right.    \nonumber   \\
    &  \quad \left. + \lambda \sum_{e=0}^{E-1} \nabla f(w_{j,i_m^K}^{e}, \xi_{j,i_m^K}^{e}) \right).
\end{IEEEeqnarray} }
where $i_{m}^{K}$ represents the index of the client with the maximum loss.
Based on Eq.~\eqref{aproof:agg-rule}, we have
{\footnotesize
\begin{IEEEeqnarray}{Cl}\label{aproof:beforeExp}
  & ||w_{j+1} - w_{j}||^2 \notag \\
  & \quad = \eta \left\| \frac{1}{K} \sum_{i=1}^{K} q_{j,i} \sum_{e=0}^{E-1} \nabla f(w_{j,i}^{e}, \xi_{j,i}^{e}) + \lambda \sum_{e=0}^{E-1} \nabla f(w_{j,i_m^K}^{e}, \xi_{j,i_m^K}^{e}) \right\|^2.      \nonumber
\end{IEEEeqnarray}
}

For convenience, we denote $A=||w_{j+1} - w_{j}||^2 / \eta^2 $.
Then, we bound the third term $A$.
{\footnotesize
\begin{IEEEeqnarray}{cl}\label{aproof:C}
  A & = \mathbb{E} \left\| \frac{1}{K} \sum_{i=1}^{K} \sum_{e=0}^{E-1} \left( q_{j,i} \nabla f(w_{j,i}^{e}, \xi_{j,i}^{e}) - q_{i}' \nabla F_i(w_{j,i}^{e})\right)  \right.   \nonumber \\
    & \quad \left. + \lambda \sum_{e=0}^{E-1} \left(\nabla f(w_{j,i_m^K}^{e}, \xi_{j,i_m^K}^{e}) - \nabla F_{i_m^K}(w_{j,i_m^K}^{e})\right) \right\|^2    \nonumber  \\
    & \quad + \mathbb{E} \left\| \frac{1}{K} \sum_{i=1}^{K} q_{i}' \sum_{e=0}^{E-1}  \nabla F_i(w_{j,i}^{e}) + \lambda \sum_{e=0}^{E-1} \nabla F_{i_m^K}(w_{j,i_m^K}^{e}) \right\|^2.
\end{IEEEeqnarray}
}

The first term of $A$ can be bounded as
{\footnotesize
\begin{IEEEeqnarray}{cl}\label{aproof:C1}
  & \mathbb{E} \left\| \frac{1}{K} \sum_{i=1}^{K} \sum_{e=0}^{E-1} \left( q_{j,i} \nabla f(w_{j,i}^{e}, \xi_{j,i}^{e}) - q_{i}' \nabla F_i(w_{j,i}^{e})\right)  \right.   \nonumber \\
    & \quad \left. + \lambda \sum_{e=0}^{E-1} \left(\nabla f(w_{j,i_m^K}^{e}, \xi_{j,i_m^K}^{e}) - \nabla F_{i_m^K}(w_{j,i_m^K}^{e})\right) \right\|^2    \nonumber  \\
  & = \mathbb{E} \left[ \frac{1}{K^2} \sum_{i=1}^{K} \sum_{e=0}^{E-1} \left\| q_{j,i} \nabla f(w_{j,i}^{e}, \xi_{j,i}^{e}) - q_{i}' \nabla F_i(w_{j,i}^{e}) \right\|^2  \right]  \nonumber \\
    & \quad  + \lambda^2  \sum_{e=0}^{E-1} \mathbb{E} \left\|\nabla f(w_{j,i_m^K}^{e}, \xi_{j,i_m^K}^{e}) - \nabla F_{i_m^K}(w_{j,i_m^K}^{e}) \right\|^2 +     \nonumber  \\
    & \quad  \sum_{e=0}^{E-1} \sum_{e=0}^{E-1} \mathbb{E} \left[ \sum_{i=1}^{K}  \frac{2 \lambda q_{i}'}{K} \left\|\nabla f(w_{j,i_m^K}^{e}, \xi_{j,i_m^K}^{e}) - \nabla F_{i_m^K}(w_{j,i_m^K}^{e}) \right\|^2  \right]    \label{ap:eq-C1}  \\
  & \leq \mathbb{E} \left[ \frac{1}{K^2} \sum_{i=1}^{K} \sum_{e=0}^{E-1} \left\| q_{j,i} \nabla f(w_{j,i}^{e}, \xi_{j,i}^{e}) - q_{j,i} \nabla F_i(w_{j,i}^{e})    \right. \right.   \nonumber  \\
    & \quad \left. \left. + q_{j,i} \nabla F_i(w_{j,i}^{e}) - q_{i}' \nabla F_i(w_{j,i}^{e}) \right\|^2  \right]  \nonumber \\
    & \quad  + \lambda^2 E G_{l}^{2}   +  \mathbb{E} \left[ \sum_{i=1}^{K}  \frac{2 \lambda q_{i}'}{K} G_{l}^2 E^2 \right]    \label{aproof:C-eq1}  \\
  & \leq \lambda^2 E G_{l}^{2}  +  \frac{2 \lambda q}{P} G_{l}^2 E^2     \nonumber \\
    & \quad  + \mathbb{E} \left[ \frac{1}{K^2} \sum_{i=1}^{K} \sum_{e=0}^{E-1} \left\| q_{j,i} \nabla f(w_{j,i}^{e}, \xi_{j,i}^{e}) - q_{j,i} \nabla F_i(w_{j,i}^{e})    \right\|^2 \right]   \nonumber  \\
    & \quad  + \mathbb{E} \left[ \frac{1}{K^2} \sum_{i=1}^{K} \sum_{e=0}^{E-1} \left\| q_{j,i} \nabla F_i(w_{j,i}^{e}) - q_{i}' \nabla F_i(w_{j,i}^{e}) \right\|^2  \right]  \nonumber \\
  & \leq \lambda^2 E G_{l}^{2}  +  \frac{2 \lambda q}{P} G_{l}^2 E^2  + \mathbb{E} \left[ \frac{1}{K^2} \sum_{i=1}^{K} E  q_{j,i}^2 G_{l}^{2} \right]  \nonumber \\
    & \quad + \mathbb{E} \left[ \frac{1}{K^2} \sum_{i=1}^{K} \sum_{e=0}^{E-1} (q_{j,i} - q_{i}')^2 \left\|  \nabla F_i(w_{j,i}^{e})  \right\|^2  \right]  \label{ap:eq-C2} \\
  & \leq \lambda^2 E G_{l}^{2}  +  \frac{2 \lambda q}{P} G_{l}^2 E^2  + \mathbb{E} \left[ \frac{1}{K^2} \sum_{i=1}^{K} E  q_{j,i}^2 G_{l}^{2} \right]  \nonumber \\
    & \quad + \mathbb{E} \left[ \frac{2}{K^2} \sum_{i=1}^{K} \sum_{e=0}^{E-1} (q_{j,i} - q_{i}')^2 \left( \left\|  \nabla F_i(w_{j,i}^{e}) - \nabla F_{i} (w_{j}) \right\|^2  \right) \right]  \nonumber   \\
    & \quad + \mathbb{E} \left[ \frac{2}{K^2} \sum_{i=1}^{K} \sum_{e=0}^{E-1} (q_{j,i} - q_{i}')^2 \left( \left\| \nabla F_{i} (w_{j})  \right\|^2  \right) \right]  \nonumber   \\
  & \leq \lambda^2 E G_{l}^{2}  +  \frac{2 \lambda q}{P} G_{l}^2 E^2  + \mathbb{E} \left[ \frac{1}{K^2} \sum_{i=1}^{K} E  q_{j,i}^2 G_{l}^{2} \right]  \nonumber \\
    & \quad + \mathbb{E} \left[ \frac{2}{K^2} \sum_{i=1}^{K} \sum_{e=0}^{E-1} (q_{j,i} - q_{i}')^2 \left( L^2 \left\|  w_{j,i}^{e} - w_{j} \right\|^2 + G_{g}^2  \right) \right].   \label{ap:eq-C3}
\end{IEEEeqnarray}
}
Eq.~\eqref{ap:eq-C1} is derived by taking expectation over the sampled data $\{\xi_{j,i}^{e}, i=1,\ldots, K$, and $e = 0, \ldots, E-1\}$. The quadratic terms in Eq.~\eqref{aproof:C-eq1}, Eq.~\eqref{ap:eq-C2}, and Eq.~\eqref{ap:eq-C3} are bounded based on the property of client-level bounded variance and global-level bounded variance as stated in Assumption~\ref{assume-local}.

Then, considering Assumption~\ref{assume-Lsmooth} and Assumption~\ref{assume-local}, we bound the second term of $C$ as follows
{\footnotesize
\begin{IEEEeqnarray}{cl}\label{aproof:C2}
   & \quad \mathbb{E}\left\| \frac{1}{K} \sum_{i=1}^{K} q_{i}' \sum_{e=0}^{E-1}  \nabla F_i(w_{j,i}^{e}) + \lambda \sum_{e=0}^{E-1} \nabla F_{i_m^K}(w_{j,i_m^K}^{e}) \right\|^2    \nonumber  \\
  & \leq 2 \mathbb{E}  \left[  \frac{E}{K} \sum_{i=1}^{K} \sum_{e=0}^{E-1} {q_{i}'}^2 \left\| \nabla F_i(w_{j,i}^{e})  \right\|^2  + \lambda^{2} E \sum_{e=0}^{E-1} \left\|  \nabla F_{i_m^K}(w_{j,i_m^K}^{e}) \right\|^2  \right] \nonumber   \\
  & \leq 4 \mathbb{E}  \left[  \frac{E}{K} \sum_{i=1}^{K} \sum_{e=0}^{E-1} {q_{i}'}^2  \left( L^2 \left\|  w_{j,i}^{e} - w_{j} \right\|^2 + G_{g}^{2}  \right)  \right]   \nonumber   \\
    & \quad + 4  \lambda^{2} E \mathbb{E} \sum_{e=0}^{E-1} \left( L^2 \left\|  w_{j,i_m^K}^{e}  -  w_{j}  \right\|^2  +  G_{g}^{2}  \right).
\end{IEEEeqnarray}
}

By applying the Eq.~\eqref{ap:eq-C3} and Eq.~\eqref{aproof:C2} to the term $A$ (Eq.~\eqref{aproof:C}), we have
{\footnotesize
\begin{IEEEeqnarray}{cl}\label{aproof:C-he}
  A & \leq \lambda^2 E G_{l}^{2}  +  \frac{2 \lambda q}{P} G_{l}^2 E^2  + \mathbb{E} \left[ \frac{1}{K^2} \sum_{i=1}^{K} E  q_{j,i}^2 G_{l}^{2} \right]  \nonumber \\
    & \quad + \mathbb{E} \left[ \frac{2}{K^2} \sum_{i=1}^{K} \sum_{e=0}^{E-1} (q_{j,i} - q_{i}')^2 \left( L^2 \left\|  w_{j,i}^{e} - w_{j} \right\|^2 + G_{g}^2  \right) \right]  \nonumber  \\
    & \quad + 4 \mathbb{E}  \left[  \frac{E}{K} \sum_{i=1}^{K} \sum_{e=0}^{E-1} {q_{i}'}^2  \left( L^2 \left\|  w_{j,i}^{e} - w_{j} \right\|^2 + G_{g}^{2}  \right)  \right]   \nonumber   \\
    & \quad + 4  \lambda^{2} E \mathbb{E} \sum_{e=0}^{E-1} \left( L^2 \left\|  w_{j,i_m^K}^{e}  -  w_{j}  \right\|^2  +  G_{g}^{2}  \right)  \nonumber  \\
  & = \lambda^2 E G_{l}^{2} + 4 G_{g}^2  \lambda^{2} E^2 +  \frac{2 \lambda q}{P} G_{l}^2 E^2  + \mathbb{E} \left[ \frac{1}{K^2} \sum_{i=1}^{K} E  q_{j,i}^2 G_{l}^{2} \right]  \nonumber \\
    & \quad + G_{g}^2 \mathbb{E} \left[ \frac{2}{K^2} \sum_{i=1}^{K} E (q_{j,i} - q_{i}')^2  + \frac{4}{K} \sum_{i=1}^{K} E^2 {q_{i}'}^2  \right]    \nonumber  \\
    & \quad + L^2 \mathbb{E} \left[ \sum_{i=1}^{K} \sum_{e=0}^{E-1}  \left( \frac{2}{K^2}  (q_{j,i} - q_{i}')^2 +  \frac{4 E {q_{i}'}^2 }{K}  \right)  \left\|  w_{j,i}^{e} - w_{j} \right\|^2   \right]  \nonumber  \\
    & \quad + 4  \lambda^{2} E \mathbb{E} \sum_{e=0}^{E-1} \left( L^2 \left\|  w_{j,i_m^K}^{e}  -  w_{j}  \right\|^2   \right).
\end{IEEEeqnarray}
}

Based on Assumption~\ref{assume-Lsmooth}, we have
{\footnotesize
\begin{IEEEeqnarray}{cl}\label{aproof:all}
  F(w_{j+1}) - F(w_{j}) & \leq \left\langle \nabla F(w_{j}), w_{j+1} - w_{j} \right\rangle \nonumber \\
  & \quad + \frac{L(1+\lambda)}{2} \| w_{j+1} - w_{j} \|^{2}.
\end{IEEEeqnarray}
}

The first term of the right side can be formulated as
{\footnotesize
\begin{IEEEeqnarray}{cl}\label{aproof:A}
  & \left\langle \nabla F(w_{j}), w_{j+1} - w_{j} \right\rangle \nonumber \\
  = &  - \eta \left\langle \nabla F(w_{j}), \frac{1}{K} \sum_{i=1}^{K} q_{j,i} \sum_{e=0}^{E-1} \nabla f(w_{j,i}^{e}, \xi_{j,i}^{e})   \right.    \nonumber   \\
    &  \left. +  \lambda \sum_{e=0}^{E-1} \nabla f(w_{j,i_m^K}^{e}, \xi_{j,i_m^K}^{e}) \right\rangle . \nonumber
\end{IEEEeqnarray}
}

By taking expectation on both sides, we bound it as follows
{\footnotesize
\begin{IEEEeqnarray}{cl}\label{aproof:A2}
  & \mathbb{E} \left\langle \nabla F(w_{j}), w_{j+1} - w_{j} \right\rangle \nonumber \\
  = &  - \eta \mathbb{E} \left\langle \nabla F(w_{j}), \frac{1}{K} \sum_{i=1}^{K} q_{j,i} \sum_{e=0}^{E-1} \nabla F_{i}(w_{j,i}^{e})   +  \lambda \sum_{e=0}^{E-1} \nabla F_{i_{m}^{K}} (w_{j,i_m^K}^{e}) \right\rangle  \nonumber   \\
  = &  - \eta \mathbb{E} \left\langle \nabla F(w_{j}), \frac{1}{K} \sum_{i=1}^{K} q_{j,i} \sum_{e=0}^{E-1} \nabla F_{i}(w_{j,i}^{e})  +  \lambda \sum_{e=0}^{E-1} \nabla F_{i_{m}^{K}} (w_{j,i_m^K}^{e}) \right.    \nonumber   \\
      &  \left. -  \frac{1}{K}\sum_{i=1}^{K} E q_{i} \nabla F_{i}(w_{j})  - \lambda E \nabla F_{i_{m}^{K}}(w_{j,i_m^K}) + E \nabla F(w_{j})  \right\rangle  \nonumber   \\
  \leq &  - \eta (E - \frac{1}{2}) \mathbb{E} \left\| \nabla F(w_{j}) \right\|^{2} + \frac{\eta}{2} \mathbb{E} \left\| \frac{1}{K} \sum_{i=1}^{K} q_{j,i} \sum_{e=0}^{E-1} \nabla F_{i}(w_{j,i}^{e}) \right.    \nonumber   \\
      &  \left. +  \lambda \sum_{e=0}^{E-1} \nabla F_{i_{m}^{K}} (w_{j,i_m^K}^{e})  -  \frac{1}{K}\sum_{i=1}^{K} E q_{i} \nabla F_{i}(w_{j})  - \lambda E \nabla F_{i_{m}^{K}}(w_{j,i_m^K})   \right\|^{2}  \nonumber   \\
  \leq &  - \eta (E - \frac{1}{2}) \mathbb{E} \left\| \nabla F(w_{j}) \right\|^{2} + \eta \mathbb{E} \left\| \frac{1}{K} \sum_{i=1}^{K} q_{j,i} \sum_{e=0}^{E-1} \nabla F_{i}(w_{j,i}^{e}) -  \frac{E}{K}  \sum_{i=1}^{K}   q_{i}   \right.    \nonumber   \\
     &  \left. \nabla F_{i}(w_{j}) \right\|^{2} + \eta \lambda^2 \mathbb{E} \left\| \sum_{e=0}^{E-1} \nabla F_{i_{m}^{K}} (w_{j,i_m^K}^{e}) -  E \nabla F_{i_{m}^{K}}(w_{j,i_m^K})   \right\|^{2}
\end{IEEEeqnarray}}
where $\nabla F(w_{j}) = \mathbb{E} \left[ \frac{1}{K}\sum_{i=1}^{K} q_{i} \nabla F_{i}(w_{j}) + \lambda \nabla F_{i_{m}^{K}}(w_{j,i_m^K}^{e}) \right] $.
We first bound the term as follows
{\footnotesize
\begin{IEEEeqnarray}{cl}\label{aproof:Asub1}
  \mathbb{E} & \left\| \frac{1}{K} \sum_{i=1}^{K} q_{j,i} \sum_{e=0}^{E-1} \nabla F_{i}(w_{j,i}^{e}) -  \frac{1}{K}\sum_{i=1}^{K} E q_{i}' \nabla F_{i}(w_{j}) \right\|^{2} \nonumber
  \\
   & \leq \frac{E}{K} \sum_{i=1}^{K} \sum_{e=0}^{E-1} \mathbb{E} \| q_{j,i} \nabla F_{i}(w_{j,i}^{e}) - q_{i}' \nabla F_{i}(w_{j}) \|^{2}  \nonumber
  \\
    & \leq \frac{2E}{K} \sum_{i=1}^{K} \sum_{e=0}^{E-1} \mathbb{E} [ (2L^{2}(q_{j,i} - q_{i}')^{2}  + q_{i}'^{2} L^{2} ) \| w_{j,i}^{e} - w_{j} \|^{2}   \nonumber   \\
    & \qquad + 4 (q_{j,i} - q_{i}')^{2}  (G_{g}^{2} + \| \nabla F(w_{j})\|^{2})  ]
\end{IEEEeqnarray}
}

Then, we have
{\footnotesize
\begin{IEEEeqnarray}{cl}\label{aproof:Asub2}
  \mathbb{E} & \left\| \lambda \sum_{e=0}^{E-1} \nabla F_{i_{m}^{K}} (w_{j,i_m^K}^{e}) - \lambda E \nabla F_{i_{m}^{K}}(w_{j,i_m^K})   \right\|^{2}   \nonumber
  \\
  & \leq \lambda^{2} E \sum_{e=0}^{E-1} \mathbb{E} \left\| \nabla F_{i_{m}^{K}} (w_{j,i_m^K}^{e}) -  \nabla F_{i_{m}^{K}}(w_{j,i_m^K})   \right\|^{2}   \nonumber
  \\
  & \leq \lambda^{2} E \sum_{e=0}^{E-1} L^{2} \mathbb{E} \left\| w_{j,i_m^K}^{e} -  w_{j,i_m^K}   \right\|^{2}
\end{IEEEeqnarray}

}

By applying the Eq.~\eqref{aproof:Asub1} and Eq.~\eqref{aproof:Asub1} to Eq.~\eqref{aproof:A2}, we have
{\footnotesize
\begin{IEEEeqnarray}{cl}\label{aproof:Ahe}
  & \mathbb{E} \left\langle \nabla F(w_{j}), w_{j+1} - w_{j} \right\rangle \nonumber \\
  & \leq  - \eta (E - \frac{1}{2}) \mathbb{E} \left\| \nabla F(w_{j}) \right\|^{2}  \nonumber  \\
  & + \frac{2\eta E}{K} \sum_{i=1}^{K} \sum_{e=0}^{E-1} \mathbb{E} [ (2L^{2}(q_{j,i} - q_{i}')^{2}  + q_{i}'^{2} L^{2} ) \| w_{j,i}^{e} - w_{j} \|^{2}   \nonumber   \\
    & \qquad + 4 (q_{j,i} - q_{i}')^{2}  (G_{g}^{2} + \| \nabla F(w_{j})\|^{2})  ]  \nonumber  \\
    & + \eta \lambda^{2} E \sum_{e=0}^{E-1} L^{2} \mathbb{E} \left\| w_{j,i_m^K}^{e} -  w_{j,i_m^K}   \right\|^{2}
\end{IEEEeqnarray}
}

By applying the upper bound of the term A in Eq.~\eqref{aproof:C-he} and Eq.~\eqref{aproof:Ahe} to the Eq.~\eqref{aproof:all}, we obtain the following expression
{\footnotesize
\begin{IEEEeqnarray}{cl}\label{aproof:main}
  & \mathbb{E} F(w_{j+1}) - F(w_{j})  \leq - \eta (E - \frac{1}{2}) \mathbb{E} \left\| \nabla F(w_{j}) \right\|^{2}  \nonumber  \\
    & + \frac{2\eta E}{K} \sum_{i=1}^{K} \sum_{e=0}^{E-1} \mathbb{E} [ (2L^{2}(q_{j,i} - q_{i}')^{2}  + q_{i}'^{2} L^{2} ) \| w_{j,i}^{e} - w_{j} \|^{2}   \nonumber   \\
    & \qquad + 4 (q_{j,i} - q_{i}')^{2}  (G_{g}^{2} + \| \nabla F(w_{j})\|^{2})  ]  \nonumber  \\
    & + \eta \lambda^{2} E \sum_{e=0}^{E-1} L^{2} \mathbb{E} \left\| w_{j,i_m^K}^{e} -  w_{j,i_m^K}   \right\|^{2} \nonumber \\
    & \quad + \frac{L(1+\lambda)}{2} \lambda^2 E G_{l}^{2} + 4 G_{g}^2  \lambda^{2} E^2 +  \frac{2 \lambda q}{P} G_{l}^2 E^2  + \mathbb{E} \left[ \frac{1}{K^2} \sum_{i=1}^{K} E  q_{j,i}^2 G_{l}^{2} \right]  \nonumber \\
    & \quad + G_{g}^2 \mathbb{E} \left[ \frac{2}{K^2} \sum_{i=1}^{K} E (q_{j,i} - q_{i}')^2  + \frac{4}{K} \sum_{i=1}^{K} E^2 {q_{i}'}^2  \right]    \nonumber  \\
    & \quad + L^2 \mathbb{E} \left[ \sum_{i=1}^{K} \sum_{e=0}^{E-1}  \left( \frac{2}{K^2}  (q_{j,i} - q_{i}')^2 +  \frac{4 E {q_{i}'}^2 }{K}  \right)  \left\|  w_{j,i}^{e} - w_{j} \right\|^2   \right]  \nonumber  \\
    & \quad + 4  \lambda^{2} E \mathbb{E} \sum_{e=0}^{E-1} \left( L^2 \left\|  w_{j,i_m^K}^{e}  -  w_{j}  \right\|^2   \right)   \nonumber \\
  & \leq  \frac{L(1+\lambda)}{2} \left( \lambda^2 E G_{l}^{2} + 4 G_{g}^2  \lambda^{2} E^2 +  \frac{2 \lambda q}{P} G_{l}^2 E^2  +  \frac{1}{K^2} \sum_{i=1}^{K} E G_{l}^{2} \mathbb{E} q_{j,i}^2  \right) \nonumber \\
    & \quad + G_{g}^2 \frac{L(1+\lambda)}{2} \mathbb{E} \left[ \frac{2}{K^2} \sum_{i=1}^{K} E (q_{j,i} - q_{i}')^2  + \frac{4}{K} \sum_{i=1}^{K} E^2 {q_{i}'}^2  \right]    \nonumber  \\
    & + \frac{2\eta E}{K} \sum_{i=1}^{K} \sum_{e=0}^{E-1} 4 (q_{j,i} - q_{i}')^{2}  G_{g}^{2}   \nonumber  \\
    &- \eta (E - \frac{1}{2}) \mathbb{E} \left\| \nabla F(w_{j}) \right\|^{2} +  \frac{2\eta E}{K} \sum_{i=1}^{K} \sum_{e=0}^{E-1}  4 (q_{j,i} - q_{i}')^{2} \mathbb{E} \left\| \nabla F(w_{j}) \right\|^{2}   \nonumber  \\
    & + \frac{2\eta E}{K} \sum_{i=1}^{K} \sum_{e=0}^{E-1} \mathbb{E} [ (2L^{2}(q_{j,i} - q_{i}')^{2}  + q_{i}'^{2} L^{2} ) \| w_{j,i}^{e} - w_{j} \|^{2} ]  \nonumber   \\
    & + \eta \lambda^{2} E \sum_{e=0}^{E-1} L^{2} \mathbb{E} \left\| w_{j,i_m^K}^{e} -  w_{j,i_m^K}   \right\|^{2} \nonumber \\
    & \quad + L^2 \frac{L(1+\lambda)}{2} \mathbb{E} \left[ \sum_{i=1}^{K} \sum_{e=0}^{E-1}  \left( \frac{2}{K^2}  (q_{j,i} - q_{i}')^2 +  \frac{4 E {q_{i}'}^2 }{K}  \right)  \left\|  w_{j,i}^{e} - w_{j} \right\|^2   \right]  \nonumber  \\
    & \quad + 4 \frac{L(1+\lambda)}{2} \lambda^{2} E \mathbb{E} \sum_{e=0}^{E-1} \left( L^2 \left\|  w_{j,i_m^K}^{e}  -  w_{j}  \right\|^2   \right) .
\end{IEEEeqnarray}
}

Then, we consider the term $\left\| w_{j,i}^{e+1} - w_{j}  \right\|^2$ in Eq.~\eqref{aproof:main}
{\footnotesize
\begin{IEEEeqnarray}{cl}
    \mathbb{E} & \left[ \left\| w_{j,i}^{e+1} - w_{j}  \right\|^2 \right]  = \mathbb{E} \left[ \left\| w_{j,i}^{e} - w_{j} -\eta \nabla f(w_{j,i}^{e}, \xi_{j,i}^{e}) \right\|^2  \right]     \nonumber   \\
    &  \leq \mathbb{E} (1+\frac{1}{K-1}) \left\| w_{j,i}^{e} - w_{j} \right\|^2 + K  \eta^2 \left\| \nabla F_i(w_{j,i}^{e}) \right\|^2  +  \eta^2 G_{l}^{2}   \nonumber   \\
    & = \mathbb{E} (1+\frac{1}{K-1}) \left\| w_{j,i}^{e} - w_{j} \right\|^2 + \eta^2 G_{l}^{2}    \nonumber   \\
      & \quad  +  K \eta^2 \left\| \nabla F_i(w_{j,i}^{e}) - \nabla F_i(w_{j}) + \nabla F_i(w_{j}) \right\|^2     \nonumber   \\
    &  \leq \mathbb{E} (1+\frac{1}{K-1}) \left\| w_{j,i}^{e} - w_{j} \right\|^2 + \eta^2 G_{l}^{2}   \nonumber   \\
      & \quad  +  2K \eta^2 \left\| \nabla F_i(w_{j,i}^{e}) - \nabla F_i(w_{j}) \right\|^2 + 2K  \eta^2 \left\|\nabla F_i(w_{j}) \right\|^2     \nonumber   \\
    &  \leq \mathbb{E} (1+\frac{1}{K-1} + 2 L^2 K  \eta^2) \left\| w_{j,i}^{e} - w_{j} \right\|^2 + \eta^2 G_{l}^{2}  +  2 K \eta^2 G_{g}^2.     \nonumber
\end{IEEEeqnarray}
}

We denote $\alpha$ as $\alpha = 1+\frac{1}{K-1} + 2 L^2 K \eta^2$. Unrolling the recursion above, we get
{\footnotesize
\begin{IEEEeqnarray}{cl}
    \left\| w_{j,i}^{e} - w_{j} \right\|^2  & \leq  \left(  \eta^2 G_{l}^2  + 2 K \eta^2 G_{g}^{2} \right) \frac{1-\alpha^e}{1-\alpha}.
\end{IEEEeqnarray}
}

By summing over $e$, we have
{\footnotesize
\begin{IEEEeqnarray}{cl}
    \sum_{e=0}^{E} \left\| w_{j,i}^{e} - w_{j} \right\|^2 & \leq  \left(  \eta^2 G_{l}^2  + 2 K \eta^2  G_{g}^{2} \right)   \left(  \frac{E-1}{1-\alpha}  - \frac{\alpha - \alpha^E}{(1-\alpha)^2}  \right). \label{ap:eq-all2}
\end{IEEEeqnarray}
}

Finally, we define $ \beta = \frac{E-1}{1-\alpha}  - \frac{\alpha - \alpha^E}{(1-\alpha)^2} $ and apply the Eq.~\eqref{ap:eq-all2} to Eq.~\eqref{aproof:main}
{\footnotesize
\begin{IEEEeqnarray}{cl}\label{aproof:main-he}
  & \mathbb{E} F(w_{j+1}) - F(w_{j})  \nonumber   \\
  & \leq  - \eta (E - \frac{1}{2}) \mathbb{E} \left\| \nabla F(w_{j}) \right\|^{2} +  \frac{2\eta E}{K} \sum_{i=1}^{K} \sum_{e=0}^{E-1}  4 (q_{j,i} - q_{i}')^{2} \mathbb{E} \left\| \nabla F(w_{j}) \right\|^{2}   \nonumber  \\
    &\quad + \frac{L(1+\lambda)}{2} \left( \lambda^2 E G_{l}^{2} + 4 G_{g}^2  \lambda^{2} E^2 +  \frac{2 \lambda q}{P} G_{l}^2 E^2  +  \frac{1}{K^2}  E G_{l}^{2}   \right) \nonumber \\
    & \quad +  \frac{G_{g}^2 L (1+\lambda)}{K}  E ( q_{\max} + 2E)  + 8 \eta E   q_{\max}  G_{g}^{2}   \nonumber  \\
    & \quad + \eta^2  \beta L^{3} (1+\lambda)  \left(   G_{l}^2  + 2 K   G_{g}^{2} \right)  \left( \frac{q_{\max} +2E }{K}  + 2 \lambda^{2} E  \right)      \nonumber  \\
    & \quad + \eta^{3} E    (4L^{2} q_{\max}  +  2L^{2} /K + \lambda^{2}  L^{2}) \left(  G_{l}^2  + 2 K   G_{g}^{2} \right) \beta     .
\end{IEEEeqnarray}

}

By simplifying both sides, we can derive the following results
\begin{IEEEeqnarray}{cl}
    & \mathcal{A} \mathbb{E} \| \nabla F(w_{j}) \|^{2}  \leq  F(w_{j}) - \mathbb{E} F(w_{j+1}) + \mathcal{B}.  \nonumber
\end{IEEEeqnarray}
where
{\footnotesize
\begin{IEEEeqnarray}{cl}
  \mathcal{A} =  &- \eta (E - \frac{1}{2}) +  \frac{8 \eta E^{2}}{K} q_{\max} ,  \label{ap:eq:final-A}
\end{IEEEeqnarray}
}
and
{\footnotesize
\begin{IEEEeqnarray}{cl}
    &  \mathcal{B} = \frac{L(1+\lambda)}{2} \left( \lambda^2 E G_{l}^{2} + 4 G_{g}^2  \lambda^{2} E^2 +  \frac{2 \lambda q}{P} G_{l}^2 E^2  +  \frac{1}{K^2}  E G_{l}^{2}   \right) \nonumber \\
    & \quad +  \frac{G_{g}^2 L (1+\lambda)}{K}  E ( q_{\max} + 2E)  + 8 \eta E   q_{\max}  G_{g}^{2}   \nonumber  \\
    & \quad + \eta^2  \beta L^{3} (1+\lambda)  \left(   G_{l}^2  + 2 K   G_{g}^{2} \right)  \left( \frac{q_{\max} +2E }{K}  + 2 \lambda^{2} E  \right)      \nonumber  \\
    & \quad + \eta^{3} E    (4L^{2} q_{\max}  +  2L^{2} /K + \lambda^{2}  L^{2}) \left(  G_{l}^2  + 2 K   G_{g}^{2} \right) \beta    .    \label{ap:eq:final-B}
\end{IEEEeqnarray}
}

\noindent where $q_{\max} \geq \sum_{i=1}^{K} \mathbb{E} [(q_{j,i} - q_{i}')^2 ]$ and $q_{\max} \leq 2$.

By accumulating both sides from $j=1$ to $J$, we get
\begin{IEEEeqnarray}{cl}
    & \frac{1}{J} \sum_{j=1}^{J} \mathcal{A} \mathbb{E} \| \nabla F(w_{j}) \|^{2}  \leq  \frac{ F(w_{1}) - F(w^*) }{J} + \mathcal{B} .  \nonumber
\end{IEEEeqnarray}

\end{IEEEproof}

\section{Additional Experiments}

\subsection{Multiple Query Attack and Defense}

\begin{figure}[!t]
  \centering
  \includegraphics[scale=0.65, trim={13 40 0 13}, clip]{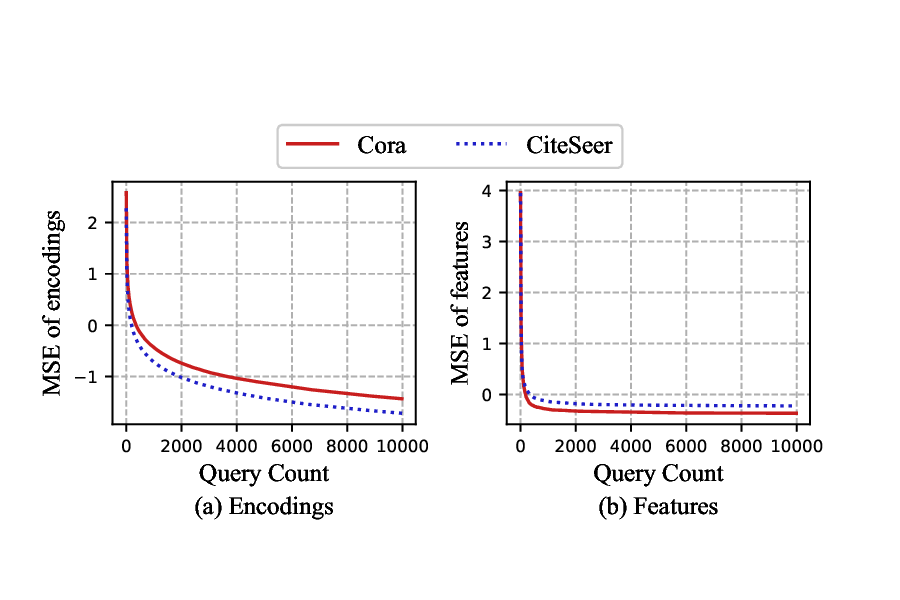}   
  \caption{Mean squared error (MSE) between the true and inferred encodings and features under multiple query attack on Cora and CiteSeer datasets. The y-axis is on a logarithmic scale. The decreasing MSE trends for both the encoding reconstruction and feature reconstruction demonstrate the increasing accuracy of the adversary.}\label{Exp:attack}
\end{figure}

\begin{table}[!ht]
  \centering
  \caption{The estimated accuracy of links under multiple query attack.}\label{ap:tab1}
  \begin{tabular}{ccccccc}
    \toprule[0.5pt]
    Query counts  &   1   &    5    &    10     &  20     & 30       \\   \midrule[0.1pt]
	Accuracy    & 72.7\%  &  87.5\% &   93.4\%  & 98.6\% &  99.6\%   \\
    \bottomrule[0.5pt]
  \end{tabular}
\end{table}

In this section, we evaluate the robustness of the proposed LDP mechanisms against the multiple query attack. This attack involves repeatedly querying a target client to upload the same perturbed outputs.

For links, the adversary can aggregate the collected results by averaging and subsequently rounding to the nearest integer. We sample 10,000 links and evaluate the estimated accuracy of this approach, as shown in Table~\ref{ap:tab1}. Experimental results demonstrate a positive correlation between the number of queries and the accuracy of link reconstruction. At 20 queries, the accuracy approaches 100\%, indicating nearly complete exposure of the raw links.

For graph nodes, the uploaded outputs are averaged to mitigate the introduced noise, with the resulting value scaled to the original encoder range before being fed into the decoder to reconstruct the sensitive raw features. In the experiment, we set the privacy budgets as $\epsilon_{a} = 3$ and $\epsilon_{b}=1$. The effectiveness of this attack was measured by the Mean Squared Error (MSE) between the true and the inferred data at both the encoding and feature reconstruction levels, as shown in Fig.~\ref{Exp:attack}.
Fig.~\ref{Exp:attack} shows a monotonic decrease in the MSE of both inferred encodings and reconstructed features as the number of adversarial queries increases. This trend indicates that the attacker can progressively remove the LDP noise through repeated sampling, leading to statistical convergence of the reconstructed data to the true values. These results confirm the multiple query attack as a realistic and potent privacy threat.

To defense these attack, our framework incorporates an effective mitigation strategy known as the permanent random response. Before training, each client noises all local data and then caches the perturbed data permanently. All subsequent queries for the same data return this identical response. This defense fundamentally invalidates the core premise of the multiple query attack. Additionally, the client uploads only a batch of noised data, making it difficult for the attacker to obtain the complete dataset.

\subsection{Impact of Encoders}
\begin{figure}[!t]
  \centering
  \includegraphics[scale=0.76, trim={13 40 0 13}, clip]{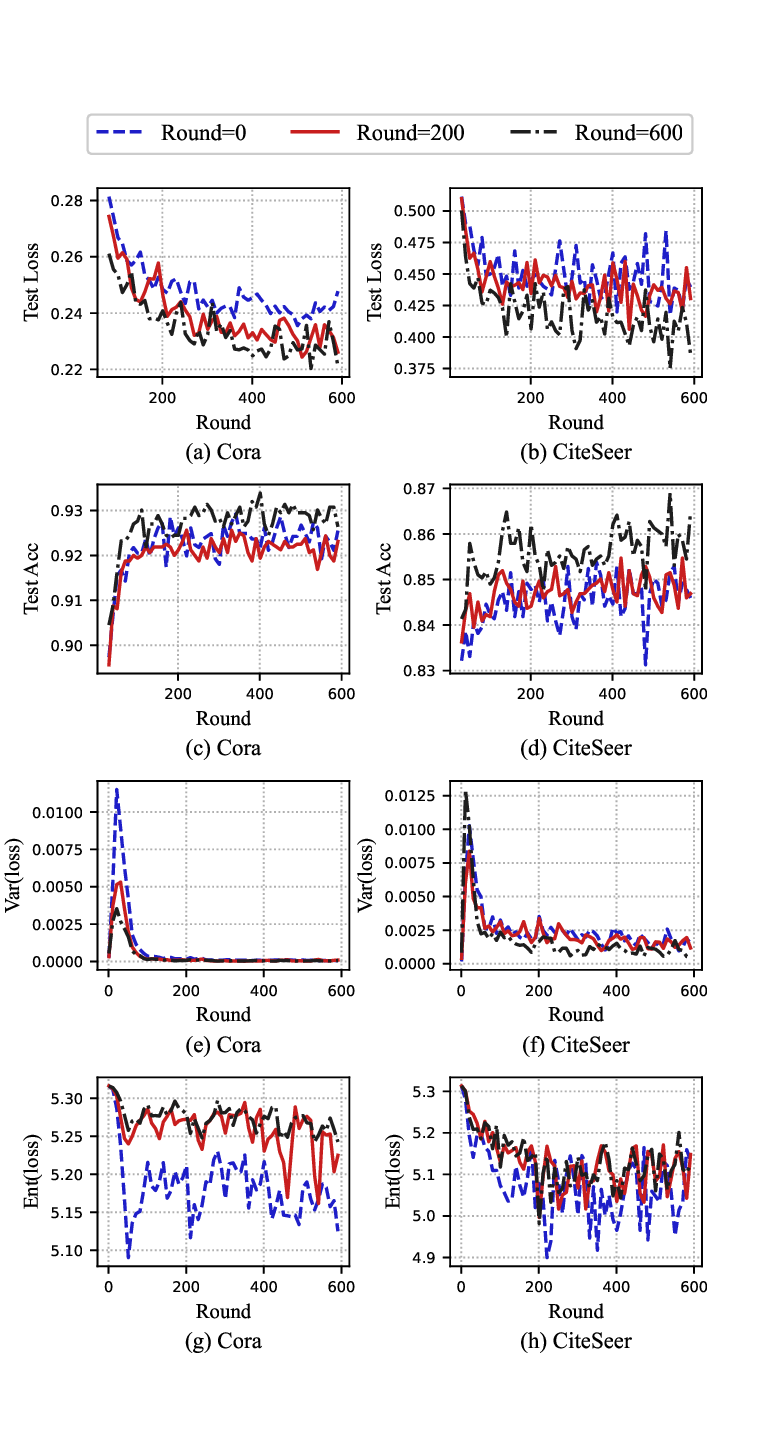}   
  \caption{The model utility and disparities (variance and entropy of local model utility) of FairGFL and the baseline under different training rounds (0,200,600) on Cora and CiteSeer.}\label{Exp:encoder}
\end{figure}

\begin{figure*}[!t]
  \centering
  \includegraphics[scale=0.7, trim={60 60 70 80}, clip]{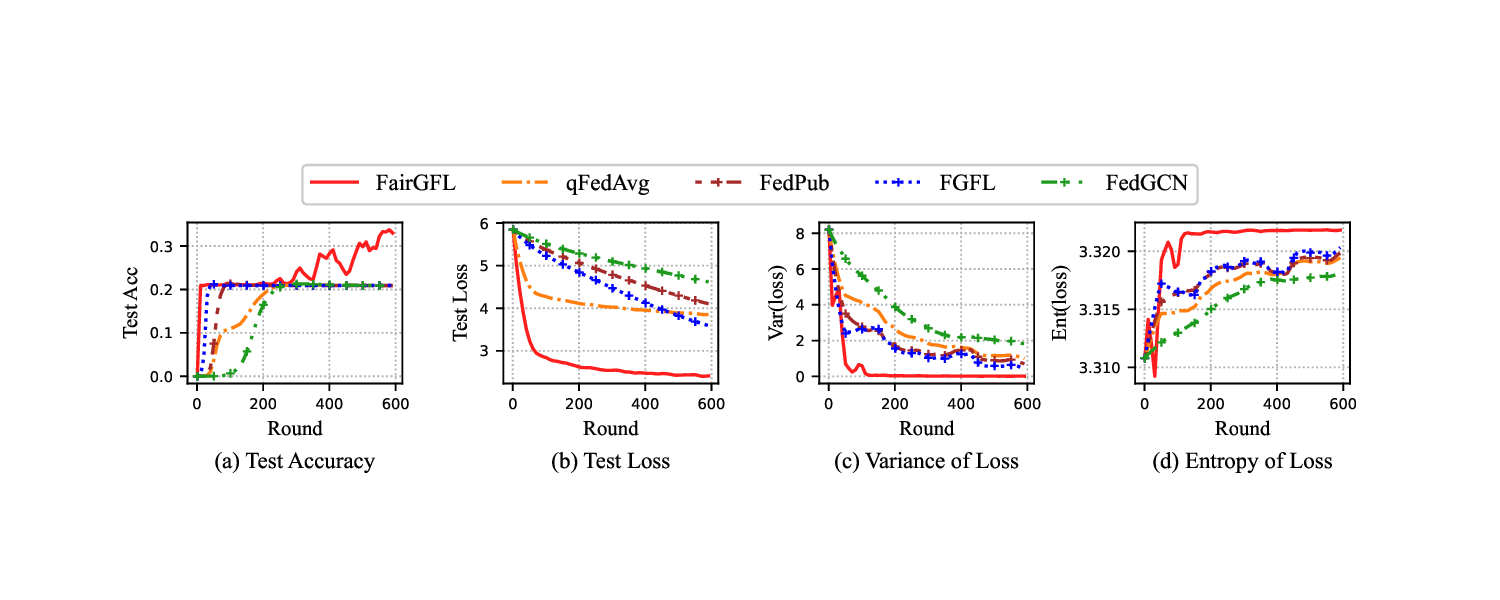}   
  \caption{The model utility and fairness (variance and entropy of local model utility) of different algorithms on ogbn-mag dataset.}\label{Exp:largedata}
\end{figure*}

\begin{figure*}[!t]
  \centering
  \includegraphics[scale=0.83, trim={55 67 56 68}, clip]{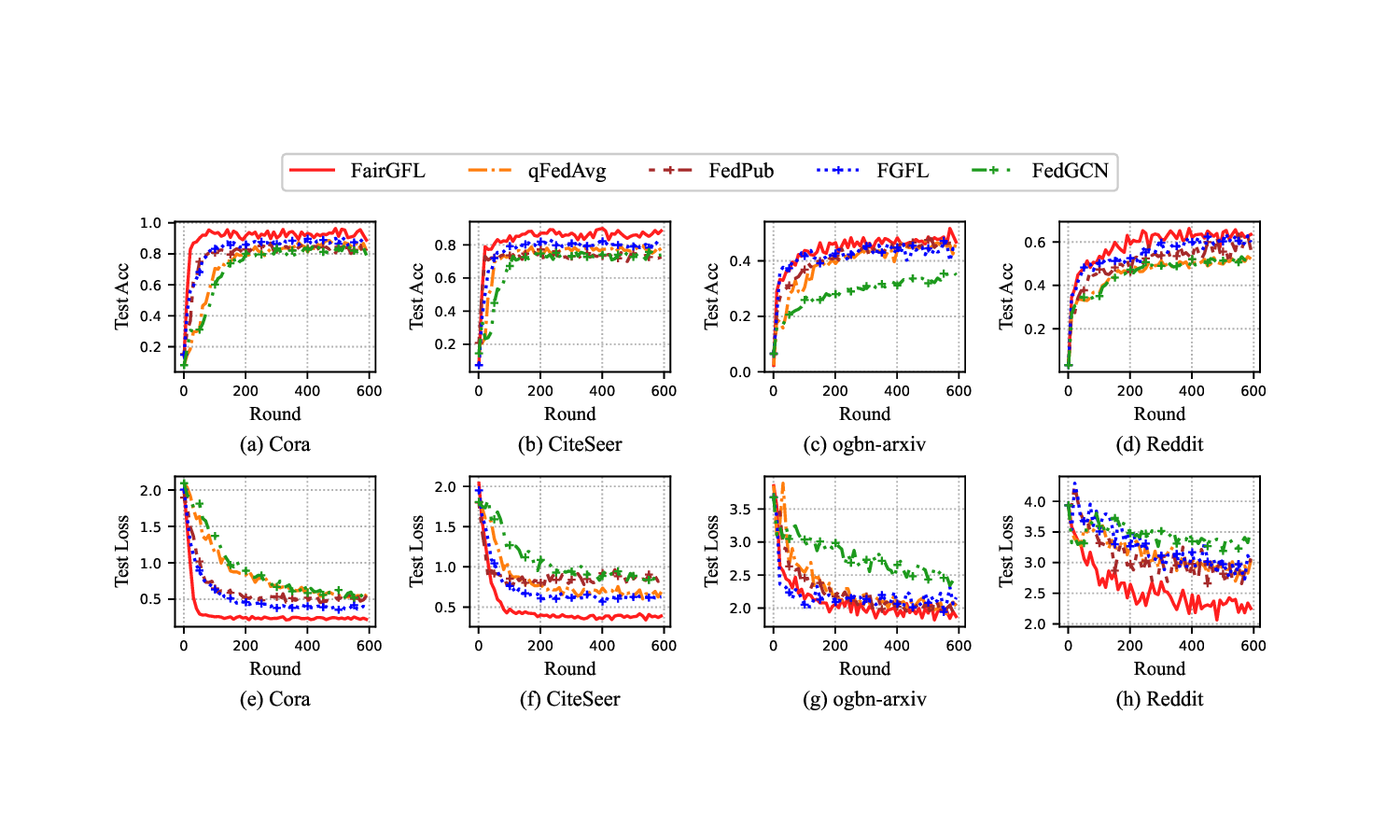}   
  \caption{The model utility of different algorithms on dynamic graph FL data (Cora, CiteSeer, Ogbn-arxiv, and Reddit). The model utility is evaluated by the test loss and accuracy.}\label{Exp:perfadynamic}
\end{figure*}

\begin{figure*}[!t]
  \centering
  \includegraphics[scale=0.83, trim={55 67 56 68}, clip]{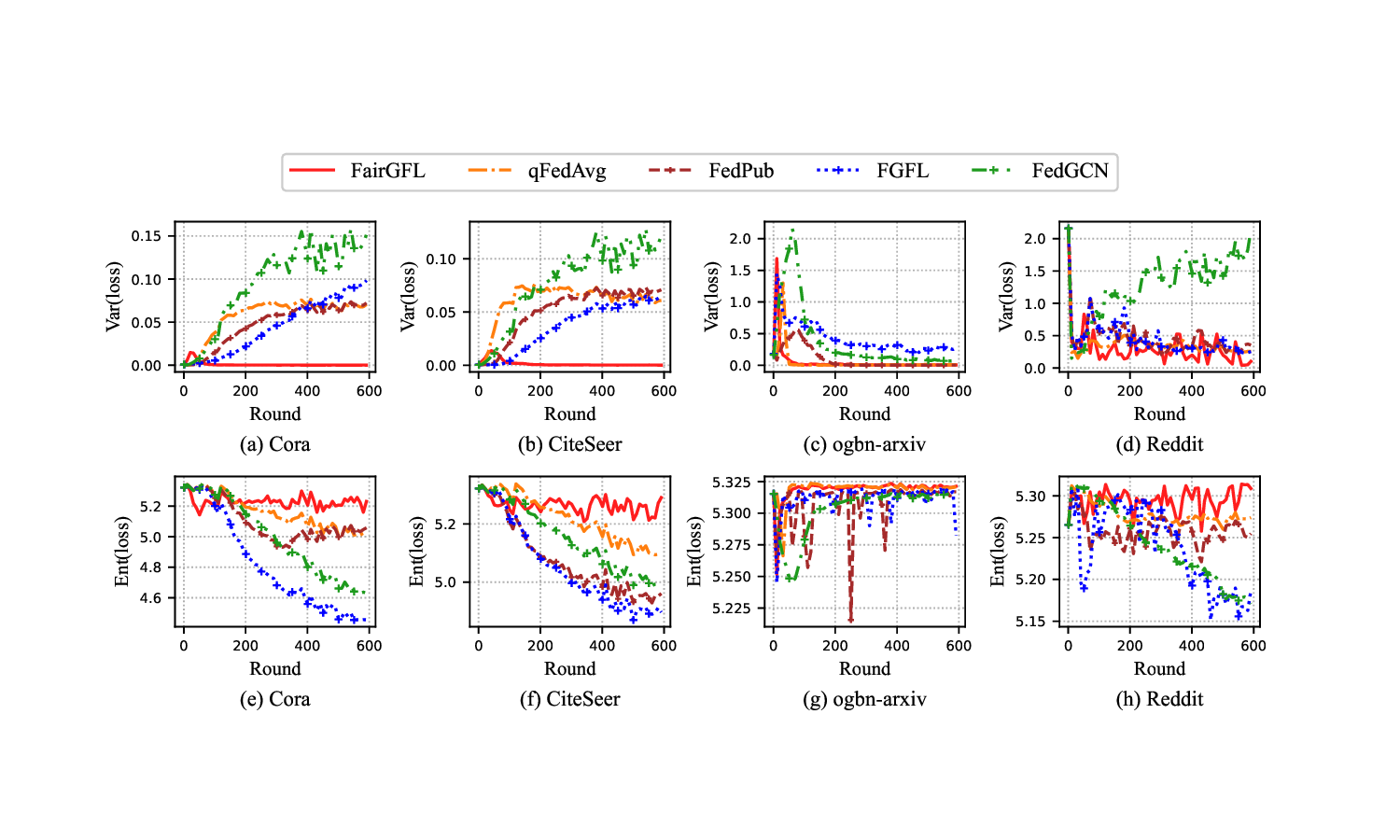}   
  \caption{The fairness (variance and entropy of local model utility) of different algorithms on dynamic graph FL data (Cora, CiteSeer, Ogbn-arxiv, and Reddit).}\label{Exp:fairadynamic}
\end{figure*}

To further investigate the impact of encoder training accuracy on model utility and fairness in FairGFL, we conducted experiments with varying autoencoder training rounds. Prior to federated training, the server collects a small number of non-privacy-sensitive nodes and trains an autoencoder on this data. Upon convergence of the autoencoder, we extract the encoder component as our feature encoding model, ensuring its capability to effectively capture essential characteristics from nodes and thereby enhance overlapping ratio estimation accuracy.

Fig.~\ref{Exp:encoder} illustrates the model performance of FairGFL under different encoder training rounds (0, 200, and 600) on both Cora and CiteSeer datasets. Experimental results demonstrate that when the autoencoder undergoes 600 training rounds, both model utility and fairness metrics achieve optimal performance. This improvement can be attributed to the encoder's sufficient training, which enables more precise extraction of distributed features from nodes. The well-trained encoder effectively reduces feature confusion and significantly improves the accuracy of overlapping ratio estimation, thereby mitigating the negative impacts of data overlaps during FL.
Conversely, insufficient training rounds (e.g., 0 or 200 rounds) result in suboptimal encoder performance. The inadequately trained encoder fails to fully capture discriminative features, potentially mapping nodes from different classes to similar latent representations. This feature ambiguity degrades the reliability of overlapping ratio estimation and leads to compromised model utility and fairness imbalance.

\subsection{Experiment Results and Analysis on Ogbn-mag}
In this section, we conducted experiments on larger dataset, ogbn-mag, to validate the robustness of FairGFL. Ogbn-mag is a heterogeneous network composed of a subset of the Microsoft Academic Graph (MAG). It is a directed graph where each node corresponds to an arXiv paper, and directed links indicate citation relationships. Each paper is associated with a 128-dimensional feature vector obtained by averaging word embeddings from its title and abstract.

Owing to the computational and memory constraints posed by the large-scale ogbn-mag dataset, we adopt a lightweight two-layer graph model, resulting in modest accuracy. However, it is not the central focus of this work. As shown in Fig.~\ref{Exp:largedata}, FairGFL nonetheless achieves superior convergence speed and fairness measured by the variance and entropy of local model utility, corroborating findings from the comparison experiments in Section~\ref{Subsec:Exp-results} and demonstrating its consistent robustness across datasets of varying sizes.

\subsection{Performance of FairGFL in Dynamic GFL}


In this section, we conducted extensive experiments under dynamic FL settings to comprehensively evaluate the robustness of FairGFL across diverse environments.

\textit{Experimental settings.}
For dynamic scenarios, we simulated realistic conditions where client data evolves gradually by constraining local data changes to 1\% per round and implementing a Dirichlet-based initial data partition. The overlapping ratio estimation mechanism was specifically adapted through reduced historical accumulation coefficients and proportional sampling from evolving data.

\textit{Experimental results and analysis.}
The results demonstrate FairGFL's consistent superiority under dynamic scenarios. Notably, in dynamic environments, the adapted overlapping estimation strategy successfully preserves accuracy of estimation despite data evolution, enabling sustained fairness and model utility. The algorithm's composite loss optimization and fairness-aware aggregation effectively handle both data heterogeneity and temporal variations, while competing methods exhibit substantial limitations: FedPub suffers from severe performance variance, FedGCN shows inadequate fairness adaptation, and FCFL displays unstable fluctuations due to its two-stage optimization. These comprehensive results validate FairGFL's robustness to environmental dynamics.

\end{document}